\documentclass{IEEEtran}

\usepackage[english]{babel}
\usepackage{amsfonts, amssymb, amsmath} 
\usepackage{bm} 
\usepackage{mathtools}
\usepackage[ruled,vlined]{algorithm2e}

\usepackage{lastpage} 
\usepackage{balance}  

\usepackage{xparse}

\usepackage{mdframed}
\mdfsetup{skipabove=0pt,skipbelow=-2pt}
\usepackage{float}

\usepackage{booktabs} 

\usepackage{tikz}
\usetikzlibrary{positioning}

\usepackage{customCommands} 

\usepackage{scalerel}
\def\dplus{\,\scalerel*{\includegraphics{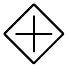}}{X\rule[-.5ex]{0pt}{1pt}\rule[1.4ex]{0pt}{1pt}}\,}
\def\dminus{\,\scalerel*{\includegraphics{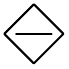}}{X\rule[-.5ex]{0pt}{1pt}\rule[1.4ex]{0pt}{1pt}}\,}

\renewcommand\Re{\operatorname{Re}}
\renewcommand\Im{\operatorname{Im}}

\newcommand{\bw}{{\bfomega}}
\newcommand{\bth}{{\bftheta}}
\newcommand{\bphi}{{\bfphi}}

\newcommand{\hhat}{^\wedge}
\newcommand{\vvee}{^\vee}

\newcommand{\mtan}[1]{T#1}
\newcommand{\mtanat}[2]{T_{#2}{#1}}

\renewcommand{\mjac}[2]{\jac{#1}{#2}} 

\newcommand{\der}{D}
\newcommand{\ndpar}[2]{\frac{\der#1}{\der#2}}
\NewDocumentCommand{\dparat}{ O{} O{} m m}{\frac{{^{#1}}\der#3}{{^{#2}}\der#4}}
\renewcommand{\rdpar}[2]{\dparat[#2][]{#1}{#2}}
\renewcommand{\ldpar}[2]{\dparat[\cE][]{#1}{#2}}
\newcommand{\rldpar}[2]{\dparat[#1][\cE]{#1}{#2}}
\newcommand{\lrdpar}[2]{\dparat[\cE][#2]{#1}{#2}}
\newcommand{\rrdpar}[2]{\dparat[#1][#2]{#1}{#2}}
\newcommand{\lldpar}[2]{\dparat[\cE][\cE]{#1}{#2}}

\newcommand{\manif}{\texttt{manif}}

\mdtheorem[nobreak=true]{example}{Example}
\floatstyle{plain}
\newfloat{float}{t}{lop}

\newenvironment{fexample}[1]
{
\begin{float}
\begin{example}[#1]
}
{
\end{example}
\end{float}
}

\usepackage[bookmarks,%
			colorlinks = true,%
			linkcolor  = black,%
			citecolor  = blue,%
			pdfauthor  = {Joan\ Sola},%
			pdftitle   = {A micro Lie theory},%
			pdftex
			]{hyperref}

\title{A micro Lie theory\\ for state estimation in robotics}
\author{Joan Sol\`a, Jeremie Deray, Dinesh Atchuthan}

\let \examples=y

\begin{document}

\maketitle


\begin{abstract}

A Lie group is an old mathematical abstract object dating back to the XIX century, when mathematician Sophus Lie laid the foundations of the theory of continuous transformation groups. 
Its influence has spread over diverse areas of science and technology many years later. 
In robotics, we are recently experiencing an important trend in its usage, at least in the fields of estimation, and particularly in motion estimation for navigation. 
Yet for a vast majority of roboticians, Lie groups are highly abstract constructions and therefore difficult to understand and to use.

In estimation for robotics it is often not necessary to exploit the full capacity of the theory, and therefore an effort of selection of materials is required. 
In this paper, we will walk through the most basic principles of the Lie theory, with the aim of conveying clear and useful ideas, and leave a significant corpus of the Lie theory behind. 
Even with this mutilation, the material included here has proven to be extremely useful in modern estimation algorithms for robotics, especially in the fields of SLAM, visual odometry, and the like. 

Alongside this micro Lie theory, we provide a chapter with a few application examples, and a vast reference of formulas for the major Lie groups used in robotics, including most Jacobian matrices and the way to easily manipulate them.
We also present a new C++ template-only library implementing all the functionality described here.

\end{abstract}




\section{Introduction}
\label{sec:intro}

There has been a remarkable effort in the last years in the robotics community to formulate estimation problems properly. 
This is motivated by an increasing demand for precision, consistency and stability of the solutions. 
Indeed, proper modeling of the states and measurements, the functions relating them, and their uncertainties, is crucial to achieving these goals.
This has led to designs involving what has been known as `manifolds', which in this context are no less than the smooth topologic surfaces of the Lie groups where the state representations evolve.
Relying on the Lie theory (LT) we are able to construct a rigorous calculus corpus to handle uncertainties, derivatives and integrals with precision and ease.
Typically, these works have focused on the well-known manifolds of rotation SO(3) and rigid motion SE(3).

\begin{figure}[tb]
\centering
\includegraphics{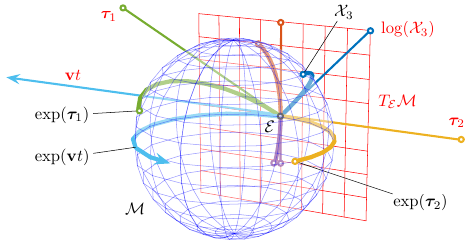}
\caption{Representation of the relation between the Lie group and the Lie algebra. 
The Lie algebra $\mtanat{\cM}{\cE}$ (red plane) is the tangent space to the Lie group's manifold $\cM$ (here represented as a blue sphere) at the identity $\cE$. 
Through the exponential map, each straight path $\bfv t$ through the origin on the Lie algebra produces a path $\exp(\bfv t)$ around the manifold which runs along the respective geodesic. 
Conversely, each element of the group has an equivalent in the Lie algebra.
This relation is so profound that (nearly) all operations in the group, which is curved and nonlinear, have an exact equivalent in the Lie algebra, which is a linear vector space.
Though the sphere in $\bbR^3$ is not a Lie group (we just use it as a representation that can be drawn on paper), that in $\bbR^4$ is, and describes the group of unit quaternions ---see \figRef{fig:manifold_q} and \exRef{ex:S3}.
}
\label{fig:exponential}
\end{figure}

When being introduced to Lie groups for the first time, it is important to try to regard them from different points of view. 
The topological viewpoint, see \figRef{fig:exponential}, involves the shape of the manifold and conveys powerful intuitions of its relation to the tangent space and the exponential map. 
The algebraic viewpoint involves the group operations and their concrete realization, allowing the exploitation of algebraic properties to develop closed-form formulas or to simplify them. 
The geometrical viewpoint, particularly useful in robotics, associates group elements to the position, velocity, orientation, and/or other modifications of bodies or reference frames. 
The origin frame may be identified with the group's identity, and any other point on the manifold represents a certain `local' frame. 
By resorting to these analogies, many mathematical abstractions of the LT can be brought closer to intuitive notions in vector spaces, geometry, kinematics, and other more classical fields.

Lie theory is by no means simple. 
To grasp a minimum idea of what LT can be, we may consider the following three references. 
First, Abbaspour's \emph{``Basic Lie theory"}~\cite{ABBASPOUR-2007-Basic_Lie_theory} comprises more than 400 pages.
With a similar title, Howe's \emph{``Very basic Lie theory"}~\cite{Howe-Basic_Lie} comprises 24 (dense) pages, and is sometimes considered a must-read introduction. 
Finally, the more modern and often celebrated Stillwell's \emph{``Naive Lie theory"}~\cite{STILLWELL-08} comprises more than 200 pages. 
With such precedents labeled as `basic', `very basic' and `naive', the aim of this paper at merely \pageref{LastPage} pages is to simplify Lie theory even more (thus our adjective `micro' in the title).
This we do in two ways. 
First, we select a small subset of material from the LT. This subset is so small that it merely explores the potential of LT. 
However, it appears very useful for uncertainty management in the kind of estimation problems we deal with in robotics (\eg~inertial pre-integration, odometry and SLAM, visual servoing, and the like), thus enabling elegant and rigorous designs of optimal optimizers.
Second, we explain it in a didactical way, with plenty of redundancy so as to reduce the entry gap to LT even more, which we believe is still needed.
That is, we insist on the efforts in this direction of, to name a paradigmatic title, Stillwell's \cite{STILLWELL-08}, and provide yet a more simplified version.
The main text body is generic, though we try to keep the abstraction level to a minimum.
Inserted examples serve as a grounding base for the general concepts when applied to known groups (rotation and motion matrices, quaternions, etc.). 
Also, plenty of figures with very verbose captions re-explain the same concepts once again. 
We put special attention to the computation of Jacobians (a topic that is not treated in \cite{STILLWELL-08}), which are essential for most optimal estimators and the source of much trouble when designing new algorithms.
We provide a chapter with some applicative examples for robot localization and mapping, implementing EKF and nonlinear optimization algorithms based on LT.
And finally, several appendices contain ample reference for the most relevant details of the most commonly used groups in robotics: unit complex numbers, quaternions, 2D and 3D rotation matrices, 2D and 3D rigid motion matrices, and the trivial translation groups.

Yet our most important simplification to Lie theory is in terms of scope. 
The following passage from 
Howe~\cite{Howe-Basic_Lie} may serve us to illustrate what we leave behind:
``\emph{The essential phenomenon of Lie theory is that one may associate in a natural way to a Lie group $\cG$ its Lie algebra $\frak{g}$. 
The Lie algebra $\frak{g}$ is first of all a vector space and secondly is endowed with a bilinear nonassociative product called the Lie bracket [...]. 
Amazingly, the group $\cG$ is almost completely determined by $\frak{g}$ and its Lie bracket. 
Thus for many purposes one can replace $\cG$ with $\frak{g}$. 
Since $\cG$ is a complicated nonlinear object and $\frak{g}$ is just a vector space, it is usually vastly simpler to work with $\frak{g}$. 
[...] 
This is one source of the power of Lie theory.%
}"
In \cite{STILLWELL-08}, Stillwell even speaks of ``\emph{the miracle of Lie theory}''.
In this work, we will effectively relegate the Lie algebra to a second plane in favor of its equivalent vector space $\bbR^n$, and will not introduce the Lie bracket at all.
Therefore, the connection between the Lie group and its Lie algebra will not be made here as profound as it should.
Our position is that, given the target application areas that we foresee, this material is often not necessary. 
Moreover, if included, then we would fail in the objective of being clear and useful, because the reader would have to go into mathematical concepts that, by their abstraction or subtleness, are unnecessarily complicated.

Our effort is in line with other recent works on the subject~\cite{BARFOOT-17-Estimation,EADE-Lie,forster2017-TRO}, which have also identified this need of bringing the LT closer to the roboticist.
Our approach aims at appearing familiar to the target audience of this paper: an audience that is skilled in state estimation (Kalman filtering, graph-based optimization, and the like), but not yet familiar with the theoretical corpus of the Lie theory.
We have for this taken some initiatives concerning notation, especially in the definition of the derivative, bringing it close to the vectorial counterparts, thus making the chain rule clearly visible.
As said, we opted to practically avoid the material proper to the Lie algebra, and prefer instead to work on its isomorphic tangent vector space $\bbR^n$, which is where we ultimately represent uncertainty or (small) state increments.
All these steps are undertaken with absolutely no loss in precision or exactness, and we believe they make the understanding of the LT and the manipulation of its tools easier.

This paper is accompanied by a new open-source C++ header-only library, called \manif\ \cite{DERAY-20-manif}, which can be found at \url{https://github.com/artivis/manif}.
\manif\ implements the widely used groups $\SO(2)$, $\SO(3)$, $\SE(2)$ and $\SE(3)$, with support for the creation of analytic Jacobians. 
The library is designed for ease of use, flexibility, and performance.


\section{A micro Lie Theory}

\subsection{The Lie group}

\begin{figure}[tb]
\centering
\includegraphics{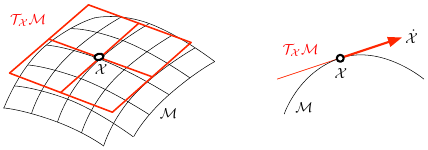}
\caption{A manifold $\cM$ and the vector space $\mtanat{\cM}{\cX}$ (in this case $\cong\bbR^2$) tangent at the point $\cX$, and a convenient side-cut.
The velocity element, $\dot\cX=\dparil{\cX}{t}$, does not belong to the manifold $\cM$ but to the tangent space $\mtanat{\cM}{\cX}$.}
\label{fig:manifold_tg}
\end{figure}

The Lie group encompasses the concepts of \emph{group} and \emph{smooth manifold} in a unique body: a Lie group $\cG$ is a smooth manifold whose elements satisfy the group axioms.
We briefly present these two concepts before joining them together.

On one hand, a differentiable or \emph{smooth manifold} is a topological space that locally resembles linear space.
The reader should be able to visualize the idea of manifold (\figRef{fig:manifold_tg}): it is like a curved, smooth (hyper)-surface, with no edges or spikes, embedded in a space of higher dimension.
In robotics, we say that our state vector evolves on this surface, that is, the manifold describes or is defined by the constraints imposed on the state.
For example, vectors with the unit norm constraint define a spherical manifold of radius one.
The smoothness of the manifold implies the existence of a unique tangent space at each point. 
This space is a linear or vector space on which we are allowed to do calculus.

On the other hand, 
a \emph{group} $(\cG,\circ)$ is a set, $\cG$, with a composition operation, $\circ$, that, for elements $\cX,\cY,\cZ\in \cG$, satisfies the following axioms, 
\begin{align}
\textrm{Closure under `$\circ$'} & ~:~~ \cX\circ \cY \in \cG  \label{equ:axiom_composition}      \\ 
\textrm{Identity $\cE$}     & ~:~~ \cE\circ \cX = \cX\circ \cE=\cX  \label{equ:axiom_identity}    \\
\textrm{Inverse $\cX\inv$}    & ~:~~ \cX\inv\circ \cX=\cX\circ \cX\inv=\cE \label{equ:axiom_inverse} \\
\textrm{Associativity}      & ~:~~ (\cX\circ \cY)\circ \cZ=\cX\circ(\cY\circ \cZ) 
~.
\end{align}

In a \emph{Lie group}, the manifold looks the same at every point (like \eg\ in the surface of a sphere, see Exs.~\ref{ex:S1} and \ref{ex:S3_intro}), and therefore all tangent spaces at any point are alike. 
The group structure imposes that the composition of elements of the manifold remains on the manifold, \eqRef{equ:axiom_composition}, and that each element has an inverse also in the manifold, \eqRef{equ:axiom_inverse}.
A special one of these elements is the identity, \eqRef{equ:axiom_identity}, and thus a special one of the tangent spaces is the tangent at the identity, which we call the Lie algebra of the Lie group. 
Lie groups join the local properties of smooth manifolds, allowing us to do calculus, with the global properties of groups, enabling the nonlinear composition of distant objects.


%
%
\begin{figure}[tb]
\centering
\includegraphics{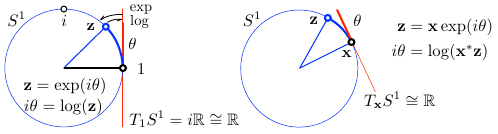}
\caption{The $S^1$ manifold is a unit circle (blue) in the plane $\bbC$, where the unit complex numbers $\bfz^*\bfz=1$ live. 
The Lie algebra $\frak{s}^1=\mtanat{S^1}{\cE}$ is the line of imaginary numbers $i\bbR$ (red), and any tangent space $\mtan{S^1}$ is isomorphic to the line $\bbR$ (red).
Tangent vectors (red segment) wrap the manifold creating the arc of circle (blue arc).
Mappings $\exp$ and $\log$ (arrows) map (wrap and unwrap) elements of $i\bbR$ to/from elements of $S^1$ (blue arc). 
Increments between unit complex numbers are expressed in the tangent space via composition and the exponential map (and we will define special operators $\op,\om$ for this).
See the text for explanations, and \figRef{fig:manifold_q} for a similar group.
}
\label{fig:manifold_z}
\end{figure}

\if\examples y

\begin{fexample}	{The unit complex numbers group $S^1$}
\label{ex:S1}

Our first example of Lie group, which is the easiest to visualize, is the group of unit complex numbers under complex multiplication (\figRef{fig:manifold_z}).
Unit complex numbers take the form $\bfz=\cos\theta+i\sin\theta$.

\emph{-- Action:}
 Vectors $\bfx=x+iy$ rotate in the plane by an angle $\theta$, through complex multiplication, $\bfx'=\bfz\,\bfx$. 

\emph{-- Group facts:}
The product of unit complex numbers is a unit complex number, the identity is 1, and the inverse is the conjugate $\bfz^*$.

\emph{-- Manifold facts:} 
The unit norm constraint defines the unit circle in the complex plane (which can be viewed as the 1-sphere, and hence the name $S^1$).
This is a  1-DoF curve in 2-dimensional space. 
Unit complex numbers evolve with time on this circle.
The group (the circle) ressembles the linear space (the tangent line) locally, but not globally.
\end{fexample}

\fi

\begin{figure}[t]
\centering
\includegraphics{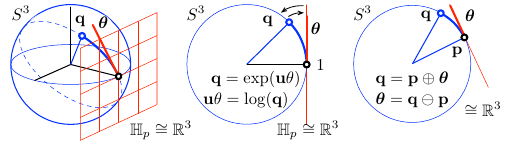}
\caption{The $S^3$ manifold is a unit 3-sphere (blue) in the 4-space of quaternions $\bbH$, where the unit quaternions $\bfq^*\,\bfq=1$ live. 
The Lie algebra is the space of pure imaginary quaternions $ix+jy+kz\in\bbH_p$, isomorphic to the hyperplane $\bbR^3$ (red grid), and any other tangent space $\mtan{S^3}$ is also isomorphic to $\bbR^3$.
Tangent vectors (red segment) wrap the manifold over the great arc or \emph{geodesic} (dashed).
The centre and right figures show a side-cut through this geodesic (notice how it resembles $S^1$ in \figRef{fig:manifold_z}).
Mappings $\exp$ and $\log$ (arrows) map (wrap and unwrap) elements of $\bbH_p$ to/from elements of $S^3$ (blue arc). 
Increments between quaternions are expressed in the tangent space via the operators $\op,\om$ (see text).
}
\label{fig:manifold_q}
\end{figure}
\if\examples y

\begin{fexample}	{The unit quaternions group $S^3$}
\label{ex:S3_intro}
A second example of Lie group, which is also relatively easy to visualize, is the group of unit quaternions under quaternion multiplication (\figRef{fig:manifold_q}).
Unit quaternions take the form $\bfq=\cos(\theta/2)+\bfu\sin(\theta/2)$, with $\bfu=iu_x+ju_y+ku_z$ a unitary axis and $\theta$ a rotation angle.

\emph{-- Action:}
Vectors $\bfx=ix+jy+kz$ rotate in 3D space  by an angle $\theta$ around the unit axis $\bfu$ through the double quaternion product $\bfx'=\bfq\,\bfx\,\bfq^*$.

\emph{-- Group facts:} 
The product of unit quaternions is a unit quaternion, the identity is 1, and the inverse is the conjugate $\bfq^*$.

\emph{-- Manifold facts:} 
The unit norm constraint defines the 3-sphere $S^3$, a spherical 3-dimensional surface or \emph{manifold} in 4-dimensional space. 
Unit quaternions evolve with time on this surface.
The group (the sphere) ressembles the linear space (the tangent hyperplane $\bbR^3\subset\bbR^4$) locally, but not globally.
\end{fexample}

\fi

\subsection{The group actions}

Importantly, Lie groups come with the power to transform elements of other sets, producing \eg~rotations, translations, scalings, and combinations of them. 
These are extensively used in robotics, both in 2D and 3D.

Given a Lie group $\cM$ and a set $\cV$, we note $\cX\cdot v$ the \emph{action} of $\cX\in\cM$ on $v\in\cV$,
\begin{align}
\cdot~:~\cM\times\cV\to\cV~;~ (\cX,v)\mapsto\cX\cdot v
~.
\end{align}
For $\cdot$ to be a group action, it must satisfy the axioms, 
\begin{align}\label{equ:action}
\textrm{Identity} &:& \cE\cdot v &= v \\
\textrm{Compatibility} &:& (\cX\circ\cY)\cdot v &= \cX\cdot(\cY\cdot v)
~.
\end{align}

Common examples are the groups of rotation matrices $\SO(n)$, the group of unit quaternions, and the groups of rigid motion $\SE(n)$. 
Their respective actions on vectors satisfy
\begin{align*}
\SO(n) &:\textrm{rotation matrix } & \bfR\cdot\bfx &\te \bfR\bfx \\
\SE(n) &:\textrm{Euclidean matrix } & \bfH\cdot\bfx &\te \bfR\bfx + \bft \\
S^1  &:\textrm{unit complex } & \bfz\cdot\bfx &\te \bfz\,\bfx \\
S^3  &:\textrm{unit quaternion } & \bfq\cdot\bfx &\te \bfq\,\bfx\,\bfq^*
\end{align*}
See \tabRef{tab:manifolds} for a more detailed exposition, and the appendices.


\begin{table*}[tb]
\caption{Typical Lie groups used in 2D and 3D motion, including the trivial $\bbR^n$. See the appendices for full reference 
}
\label{tab:manifolds}
\begin{center}
\begin{tabular}{|c|c|c|c|c|c|c|c|c|c|c|}
\multicolumn{2}{|c|}{Lie group $\cM,\circ$} & \!size\! & \!dim\! 
  & $\cX\in\cM$ 
  & Constraint
  & $\bftau^\wedge\in\frak{m}$    
  & $\bftau\in\bbR^m$ 
  & $\Exp(\bftau)$  
  & Comp.
  & Action
  \\
\toprule
$n$-D vector  & $\bbR^n,+$ & $n$  & $n$ 
  & $\bfv\in\bbR^n$ 
  & $\bfv-\bfv=\bf0$
  & $\bfv\in\bbR^n$     
  & $\bfv\in\bbR^n$  
  & $\bfv=\exp(\bfv)$ 
  & $\bfv_1\!+\!\bfv_2$
  & $\bfv + \bfx$
  \\
\midrule
circle        & $S^1,\cdot$   & 2    & 1 
  & $\bfz\in\bbC$ 
  & $\bfz^*\bfz=1$
  & $i\theta\in i\bbR$  
  & $\theta\in\bbR$  
  & $\bfz=\exp(i\theta)$ 
  & $\bfz_1\,\bfz_2$
  & $\bfz\,\bfx$
  \\
Rotation   & $\SO(2),\cdot$ & 4    & 1 
  & $\bfR$ 
  & $\bfR\tr\bfR=\bfI$
  & $\hatx{\theta}\in\so(2)$     
  & $\theta\in\bbR$    
  & $\bfR=\exp(\hatx{\theta})$ 
  & $\bfR_1\,\bfR_2$
  & $\bfR\,\bfx$
  \\
Rigid motion  
  & $\SE(2),\cdot$ & 9    & 3   
  & $\bfM=\begin{bsmallmatrix}\bfR & \bft \\ 0 & 1\end{bsmallmatrix}$ 
  & $\bfR\tr\bfR=\bfI$
  & $\begin{bsmallmatrix}\hatx{\theta} & \bfrho \\ 0 & 0 \end{bsmallmatrix} \!\in\!\se(2)$     
  & $\begin{bsmallmatrix}\bfrho\\ \theta\end{bsmallmatrix}\in\bbR^3$  
  & $\exp\left(\begin{bsmallmatrix}\hatx{\theta} & \bfrho \\ 0 & 0\end{bsmallmatrix}\right)$ 
  & $\bfM_1\,\bfM_2$
  & $\bfR\,\bfx\!+\!\bft$
  \\
\midrule
3-sphere      & $S^3,\cdot$ & 4    & 3   
  & $\bfq\in\bbH$ 
  & $\bfq^*\bfq=1$
  & $\bth/2\in\bbH_p$  
  & $\bth\in\bbR^3$  
  & $\bfq=\exp(\bfu\theta/2)$ 
  & $\bfq_1\,\bfq_2$
  & $\bfq\,\bfx\,\bfq^*$ 
  \\
Rotation   & $\SO(3),\cdot$ & 9    & 3   
  & $\bfR$ 
  & $\bfR\tr\bfR=\bfI$
  & $\hatx{\bth}\in\so(3)$     
  & $\bth\in\bbR^3$  
  & $\bfR=\exp(\hatx{\bth})$ 
  & $\bfR_1\,\bfR_2$
  & $\bfR\,\bfx$
  \\
Rigid motion  & $\SE(3),\cdot$   & 16   & 6 
  & $\bfM=\begin{bsmallmatrix}\bfR & \bft \\ 0 & 1\end{bsmallmatrix}$ 
  & $\bfR\tr\bfR=\bfI$
  & $\begin{bsmallmatrix}\hatx{\bth} & \bfrho \\ 0 & 0\end{bsmallmatrix} \!\in\!\se(3)$     
  & $\begin{bsmallmatrix}\bfrho\\\bth\end{bsmallmatrix}\in\bbR^6$  
  & $\exp\left(\begin{bsmallmatrix}\hatx{\bth} & \bfrho \\ 0 & 0\end{bsmallmatrix}\right)$ 
  & $\bfM_1\,\bfM_2$
  & $\bfR\,\bfx\!+\!\bft$
  \\
\bottomrule
\end{tabular}
\end{center}
\end{table*}%

The group composition \eqRef{equ:axiom_composition} may be viewed as an action of the group on itself, $\circ:\cM\times\cM\to\cM$.
Another interesting action is the \emph{adjoint action}, which we will see in \secRef{sec:adjoint}.

\subsection{The tangent spaces and the Lie algebra}

Given $\cX(t)$ a point moving on a Lie group's manifold $\cM$, its velocity $\dot\cX=\dparil{\cX}{t}$ belongs to the space tangent to $\cM$ at $\cX$ (\figRef{fig:manifold_tg}),
which we note $\mtanat{\cM}{\cX}$.
The smoothness of the manifold, \ie, the absence of edges or spikes, implies the existence of a unique tangent space at each point.
The structure of such tangent spaces is the same everywhere.

\subsubsection[The Lie algebra]{The Lie algebra $\frak{m}$}

The tangent space at the identity, $\mtanat{\cM}{\cE}$, is called the \emph{Lie algebra} of $\cM$, and noted $\frak{m}$,
\begin{align}
\textrm{Lie algebra}&:&\frak{m} &\te \mtanat{\cM}{\cE}~.
\end{align}
Every Lie group has an associated Lie algebra.
We relate the Lie group with its Lie algebra through the following facts~\cite{EADE-Lie} (see Figs.~\ref{fig:exponential} and \ref{fig:maps}):
\begin{itemize}
\item
The Lie algebra $\frak{m}$ is a vector space.\footnotemark\ 
As such, its elements can be \emph{identified} with vectors in $\bbR^m$, whose dimension $m$ is the number of degrees of freedom of $\cM$.
\footnotetext{%
In any Lie algebra, the vector space is endowed with a non-associative product called the Lie bracket. In this work, we will not make use of it.}
\item
The \emph{exponential map}, $\exp:\frak{m}\to\cM$, exactly converts elements of the Lie algebra into elements of the group. The log map is the inverse operation.
\item
Vectors of the tangent space at $\cX$ can be transformed to the tangent space at the identity $\cE$ 
through a linear transform. This transform is called the \emph{adjoint}.
\end{itemize}

\begin{figure}[tb]
\centering
\includegraphics{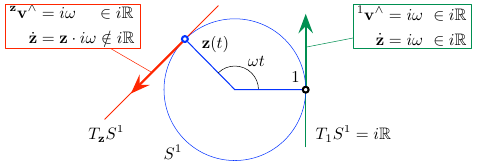}
\caption{%
Let a point $\bfz\in S^1$ move at constant rotation rate $\omega$, $\bfz(t)=\cos\omega t+i\sin\omega t$.
Its velocities when passing through $1$ and $\bfz$ are in the respective tangent spaces, $\mtanat{S^1}{1}$ and $\mtanat{S^1}{\bfz}$. 
In the case of $\mtanat{S^1}{\bfz}$, the velocity is $\dot\bfz=\bfz \,i\omega = -\omega\sin\omega t + i\omega\cos\omega t$ when expressed in the global coordinates, and ${^\bfz}\!\bfv\hhat=i\omega$ when expressed locally. 
Their relation is given by ${^\bfz}\!\bfv\hhat=\bfz\inv\dot\bfz=\bfz^*\dot\bfz$.
In the case of $\mtanat{S^1}{1}$, this relation is the identity ${^1}\!\bfv\hhat=\dot\bfz=i\omega$.
Clearly, the structure of all tangent spaces is $i\bbR$, which is the Lie algebra.
This is also the structure of $\dot\bfz$ at the identity, and this is why the Lie algebra is defined as the tangent space at the identity.
}
\label{fig:global_local_tangent}
\end{figure}

Lie algebras can be defined locally to a tangent point $\cX$, establishing local coordinates for $\mtanat{\cM}{\cX}$
(\figRef{fig:global_local_tangent}). 
We shall denote elements of the Lie algebras with a `hat' decorator, such as $\bfv\hhat$ for velocities or $\bftau\hhat=(\bfv t)\hhat=\bfv\hhat t$ for general elements. 
A left superscript may also be added to specify the precise tangent space, \eg, ${^\cX}\!\bfv\hhat\in\mtanat{\cM}{\cX}$ and ${^\cE}\!\bfv\hhat\in\mtanat{\cM}{\cE}$.

The structure of the Lie algebra can be found%
\if \examples y (see Examples~\ref{ex:SO3} and \ref{ex:S3}) \else \fi 
by time-differentiating the group constraint~\eqRef{equ:axiom_inverse}.
For multiplicative groups this yields the new constraint $\cX\inv\dot\cX + \dot{\cX\inv}\cX = 0$, which applies to the elements tangent at $\cX$ (the term $\dot{\cX\inv}$ is the derivative of the inverse). 
The elements of the Lie algebra are therefore of the form,\footnote{For additive Lie groups the constraint $\cX-\cX=0$ differentiates to $\dot\cX=\dot\cX$, that is, no constraint affects the tangent space. This means that the tangent space is the same as the group space. See \appRef{sec:Tn} for more details.}
\begin{align}\label{equ:tangent_structure}
\bfv\hhat = \cX\inv\dot\cX &= -\dot{\cX\inv}\cX
~.
\end{align}
%

\subsubsection[The Cartesian vector space]{The Cartesian vector space $\bbR^m$}

\begin{figure}[tb]
\centering
\includegraphics{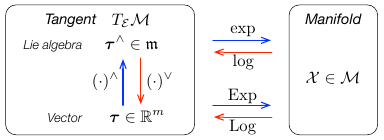}
\caption{Mappings between the manifold $\cM$ and the representations of its tangent space at the origin $\mtanat{\cM}{\cE}$ (Lie algebra $\frak{m}$ and Cartesian $\bbR^m$).
Maps hat $(\cdot)\hhat$ and vee $(\cdot)\vvee$ are the linear invertible maps or \emph{isomorphisms} \eqsRef{equ:hat}{equ:vee}, $\exp(\cdot)$ and $\log(\cdot)$ map the Lie algebra to/from the manifold, and $\Exp(\cdot)$ and $\Log(\cdot)$ are shortcuts to map directly the vector space $\bbR^m$ to/from $\cM$.}%
\label{fig:maps}%
\end{figure}%

The elements $\bftau\hhat$ of the Lie algebra have non-trivial structures (skew-symmetric matrices, imaginary numbers, pure quaternions, see \tabRef{tab:manifolds})
but the key aspect for us is that they can be expressed as linear combinations of some base elements $E_i$, where $E_i$ are called the \emph{generators} of $\frak{m}$ 
(they are the derivatives of $\cX$ around the origin in the $i$-th direction).
It is then handy to manipulate just the coordinates as vectors in $\bbR^m$, which we shall note simply $\bftau$.
We may pass from $\frak{m}$ to $\bbR^m$ and vice versa through two mutually inverse linear maps or \emph{isomorphisms}, commonly called \emph{hat} and \emph{vee} (see \figRef{fig:maps}),
\begin{align} 
\textrm{Hat}&:& 
\bbR^m&\to\frak{m}\,; 
 & 
 \bftau\,
 &\mapsto \bftau\hhat 
 = \sum_{i=1}^m \tau_i\, E_i \label{equ:hat} 
\\
\textrm{Vee}&:& 
 \frak{m}&\to\bbR^m\,; 
 & \bftau\hhat
 & \mapsto (\bftau\hhat)\vvee
 =\bftau
 = \sum_{i=1}^m \tau_i\,\bfe_i  \label{equ:vee}
~,
\end{align}
with $\bfe_i$ the vectors of the base of $\bbR^m$ (we have $\bfe_i\hhat=E_i$).
This means that $\frak{m}$ is isomorphic to the vector space $\bbR^m$ 
--- one writes $\frak{m}\cong\bbR^m$, or $\bftau\hhat\cong\bftau$.
Vectors $\bftau\in\bbR^m$ are handier for our purposes than their isomorphic $\bftau\hhat\in\frak{m}$, since they can be stacked in larger state vectors, and more importantly, manipulated with linear algebra using matrix operators. 
In this work, we enforce this preference of $\bbR^m$ over $\frak{m}$, to the point that most of the operators and objects that we define (specifically: the adjoint, the Jacobians, the perturbations and their covariances matrices, as we will see soon) are on $\bbR^m$.

\if\examples y

\begin{fexample}{The rotation group $\SO(3)$, its Lie algebra $\so(3)$, and the vector space $\bbR^3$}
\label{ex:SO3}
In the rotation group $\SO(3)$, of $3\tcross3$ rotation matrices $\bfR$, we have the orthogonality condition $\bfR\tr\bfR=\bfI$.
The tangent space may be found by taking the time derivative of this constraint, that is $\bfR\tr\dot\bfR+\dot\bfR\tr\bfR=0$, which we rearrange as $$\bfR\tr\dot\bfR=-(\bfR\tr\dot\bfR)\tr.$$
This expression reveals that $\bfR\tr\dot\bfR$ is a skew-symmetric matrix (the negative of its transpose). 
Skew-symmetric matrices are often noted $\hatx{\bw}$ and have the form 
$$\hatx{\bw}=\begin{bsmallmatrix}
0 & -\omega_z & \omega_y \\
\omega_z & 0 & -\omega_x \\
-\omega_y & \omega_x & 0 
\end{bsmallmatrix}
.$$
This gives $\bfR\tr\dot\bfR=\hatx{\bw}$. When $\bfR=\bfI$ we have $$\dot\bfR=\hatx{\bw},$$ that is, $\hatx{\bw}$ is in the Lie algebra of $\SO(3)$, which we name $\so(3)$.
Since $\hatx{\bw}\in\so(3)$ has 3 DoF, the dimension of $\SO(3)$ is $m=3$.
The Lie algebra is a vector space whose elements can be decomposed into
$$
\hatx{\bw} = 
  \omega_x\bfE_x+
  \omega_y\bfE_y+
  \omega_z\bfE_z
$$
with  
$
\bfE_x=
\begin{bsmallmatrix}0&0&0\\0&0&-1\\0&1&0\end{bsmallmatrix}$, 
$\bfE_y=
\begin{bsmallmatrix}0&0&1\\0&0&0\\-1&0&0\end{bsmallmatrix}$,
 $\bfE_z=
\begin{bsmallmatrix}0&-1&0\\1&0&0\\0&0&0\end{bsmallmatrix}$ 
the generators of $\so(3)$, and where $\bw=(\omega_x,\omega_y,\omega_z)\in\bbR^3$ is the vector of angular velocities. 
The one-to-one linear relation above allows us to identify $\so(3)$ with $\bbR^3$ --- we write $\so(3)\cong\bbR^3$.
We pass from $\so(3)$ to $\bbR^3$ and viceversa using the linear operators \emph{hat} and \emph{vee},
\begin{align*}
\textrm{Hat}&:& \bbR^3&\to\so(3);& \bw&\mapsto\bw^\wedge=\hatx{\bw} 
\\
\textrm{Vee}&:& \so(3)&\to\bbR^3;& \hatx{\bw}&\mapsto\hatx{\bw}^\vee=\bw
~.
\end{align*}
\end{fexample}

\fi

\subsection{The exponential map}

\if\examples y

\begin{fexample}{The exponential map of $\SO(3)$}
\label{ex:SO3_exp}
%
%
We have seen in \exRef{ex:SO3} that 
$
\dot\bfR = \bfR\hatx{\bfomega}  \in \mtanat{\SO(3)}{\bfR}.
$
For $\bw$ constant, this is an ordinary differential equation (ODE), whose solution is $\bfR(t) = \bfR_0\exp(\hatx{\bfomega}t)$. At the origin $\bfR_0=\bfI$ we have the exponential map,
\begin{align*}
\bfR(t) &= \exp(\hatx{\bfomega}t) && \in\SO(3)
~. 
\end{align*}
We now define the vector $\bth\te\bfu\theta \te \bfomega t\in\bbR^3$ as the integrated rotation in angle-axis form, with angle $\theta$ and unit axis $\bfu$. 
Thus $\hatx{\bth}\in\so(3)$ is the total rotation expressed in the Lie algebra.
We substitute it above. 
Then write the exponential as a power series,
\begin{align*}
\bfR &= \exp(\hatx{\bth})= \sum_k\frac{\theta^k}{k!}{(\hatx{\bfu})^k} 
~.
\end{align*}
In order to find a closed-form expression, we write down a few powers of $\hatx{\bfu}$,
\begin{align*}
\hatx{\bfu}^0&= \bfI,
&
\hatx{\bfu}^1&= \hatx{\bfu},
\\
\hatx{\bfu}^2&= \bfu\bfu\tr -\bfI,
&
\hatx{\bfu}^3&=-\hatx{\bfu},
\\
\hatx{\bfu}^4&=-\hatx{\bfu}^2,
&
\cdots
\end{align*}
and realize that all can be expressed as multiples of $\bfI$, $\hatx{\bfu}$ or $\hatx{\bfu}^2$.
We thus rewrite the series as,
%
\begin{align*}
\bfR = \bfI &+ \hatx{\bfu}\big(\theta - \tfrac1{3!}\theta^3 + \tfrac1{5!}\theta^5 - \cdots\big) \\
  &+ \hatx{\bfu}^2\big(\tfrac12\theta^2-\tfrac1{4!}\theta^4+\tfrac1{6!}\theta^6-\cdots\big)
  ~,
\end{align*}
where we identify the series of $\sin\theta$ and $\cos\theta$, yielding the closed form,
\begin{align*}
\bfR=\exp(\hatx{\bfu\theta}) &= 
\bfI + \hatx{\bfu}\sin\theta + \hatx{\bfu}^2(1\!-\!\cos\theta)
~.
\end{align*}
This expression is the well known Rodrigues rotation formula.
It can be used as the capitalized exponential just by doing $\bfR=\Exp(\bfu\theta)=\exp(\hatx{\bfu\theta})$.
\end{fexample}

\fi

The exponential map $\exp()$ allows us to exactly transfer elements of the Lie algebra to the group (\figRef{fig:exponential}), an operation generically known as \emph{retraction}. 
Intuitively, $\exp()$ wraps the tangent element around the manifold following the great arc or \emph{geodesic} (as when wrapping a string around a ball, Figs.~\ref{fig:exponential}, \ref{fig:manifold_z} and \ref{fig:manifold_q}).
The inverse map is the $\log()$, \ie, the unwrapping operation.
The $\exp()$ map arises naturally by considering the time-derivatives of $\cX\in\cM$ over the manifold, as follows.
From \eqRef{equ:tangent_structure} we have,
\begin{align}\label{equ:ode}
\dot\cX &= \cX\bfv\hhat 
~.
\end{align}
For $\bfv$ constant, this is an ordinary differential equation (ODE) whose solution is 
\begin{align}\label{equ:exp_at_X}
\cX(t)  = \cX(0)\exp(\bfv\hhat t)
~.
\end{align}
Since $\cX(t)$ and $\cX(0)$ are elements of the group, then $\exp(\bfv\hhat t)=\cX(0)\inv\cX(t)$ must be in the group too, and so $\exp(\bfv\hhat t)$ maps elements $\bfv\hhat t$ of the Lie algebra to the group.
This is known as the \emph{exponential map}.

In order to provide a more generic definition of the exponential map, let us define the tangent increment $\bftau\te\bfv t\in\bbR^m$ as velocity per time, so that we have $\bftau\hhat=\bfv\hhat t\in\frak{m}$ a point in the Lie algebra. 
The exponential map, and its inverse the logarithmic map, can be now written as, 
\begin{align}
\exp &:& \frak{m} & \to\cM      
 &&; & \bftau\hhat &\mapsto \,\cX = \exp(\bftau\hhat) 
 \\
\log &:&     \cM & \to\frak{m} 
 &&; & \cX\,    &\mapsto \bftau\hhat = \log(\cX) ~
~. 
\end{align}

Closed forms of the exponential in multiplicative groups are obtained by writing the absolutely convergent Taylor series, 
\begin{align}
\exp(\bftau\hhat)=\cE+\bftau\hhat+\tfrac12{\bftau\hhat}^2+\tfrac1{3!}{\bftau\hhat}^3+\cdots
~,
\end{align}
and taking advantage of the algebraic properties of the powers of $\bftau\hhat$%
\if \examples y (see Ex.~\ref{ex:SO3_exp} and \ref{ex:S3} for developments of the exponential map in $\SO(3)$ and $S^3$). \else. \fi 
These are then inverted to find the logarithmic map.
Key properties of the exponential map are 
\begin{align}
\exp((t+s)\bftau\hhat) 
 &= \exp(t\bftau\hhat)\exp(s\bftau\hhat) \label{equ:prop_exp_ts}
 \\
\exp(t\bftau\hhat) 
 &= \exp(\bftau\hhat)^t \label{equ:prop_exp_t}
 \\
\exp(-\bftau\hhat) 
 &= \exp(\bftau\hhat)\inv \label{equ:prop_exp_inv}
 \\
\exp(\cX\bftau\hhat\cX\inv) 
 &= \cX\exp(\bftau\hhat)\cX\inv ~, \label{equ:prop_exp}
\end{align} 
where \eqRef{equ:prop_exp}, a surprising and powerful statement, can be proved easily by expanding the Taylor series and simplifying the many terms $\cX\inv\cX$.

\if\examples y

\begin{fexample}{The unit quaternions group $S^3$ (cont.)}
\label{ex:S3}
In the  group $S^3$ (recall \exRef{ex:S3_intro} and see \eg\ \cite{SOLA-17-Quaternion}), the time derivative of the unit norm condition $\bfq^*\bfq=1$ yields 
$$\bfq^*\dot\bfq=-(\bfq^*\dot\bfq)^*
.
$$
This reveals that $\bfq^*\dot\bfq$ is a pure quaternion (its real part is zero). 
Pure quaternions $\bfu v\in\bbH_p$ have the form 
$$\bfu v=(iu_x+ju_y+ku_z)v =iv_x+jv_y+kv_z 
,$$
where $\bfu\te iu_x+ju_y+ku_z$ is pure and unitary, $v$ is the norm, and $i,j,k$ are the generators of the Lie algebra $\frak{s}^3=\bbH_p$.
Re-writing the condition above we have,
\begin{align*}
\dot\bfq=\bfq\, \bfu v &&\in \mtanat{S^3}{\bfq}
,
\end{align*}
which integrates to $\bfq = \bfq_0\exp(\bfu v t)$. 
Letting $\bfq_0=1$ and defining $\bphi \te \bfu\phi \te\bfu v t$ we get the exponential map,
\begin{align*}
\bfq = \exp(\bfu\phi) \te \sum \frac{\phi^k}{k!}\bfu^k &&\in S^3
~.
\end{align*}
The powers of $\bfu$ follow the pattern $1,\bfu,-1,-\bfu,1,\cdots$.
Thus we group the terms in $1$ and $\bfu$ and identify the series of $\cos\phi$ and $\sin\phi$.
We get the closed form,
\begin{align*}
\bfq = \exp(\bfu\phi) = \cos(\phi) + \bfu\sin(\phi) 
~,
\end{align*}
which is a beautiful extension of the Euler formula, $\exp(i\phi)=\cos\phi+i\sin\phi$.
The elements of the Lie algebra $\bfphi=\bfu \phi\in\frak{s}^3$ can be identified with the rotation vector $\bth\in\bbR^3$ trough the mappings \emph{hat} and \emph{vee},
\begin{align*}
\textrm{Hat}&:& \bbR^3&\to\frak{s}^3;& \bth&\mapsto\bth^\wedge=\bth/2
\\
\textrm{Vee}&:& \frak{s}^3&\to\bbR^3;& \bphi&\mapsto\bphi^\vee=2\bphi
~,
\end{align*}
where 
the factor 2 accounts for the double effect of the quaternion in the rotation action, $\bfx'=\bfq\,\bfx\,\bfq^*$. 
With this choice of Hat and Vee, the quaternion exponential
\begin{align*}
\bfq = \Exp(\bfu\theta)=\cos(\theta/2)+\bfu\sin(\theta/2)
\end{align*}
is equivalent to the rotation matrix $\bfR=\Exp(\bfu\theta)$.
\end{fexample}

\fi

\subsubsection[The capitalized Exp map]{The capitalized exponential map}

The capitalized Exp and Log maps are convenient shortcuts to map vector elements $\bftau\in\bbR^m~(\cong\mtanat{\cM}{\cE})$ directly with elements $\cX\in\cM$.
We have,
\begin{align}
\Exp &: &\bbR^m &\to\cM &&;&
\,\bftau &\mapsto \cX = \Exp(\bftau)\\
\Log  &: & \cM &\to\bbR^m &&;&
 \cX &\mapsto \bftau\,=\Log(\cX) 
\,. 
\end{align}
Clearly from \figRef{fig:maps},
\begin{align}
\cX  = \Exp(\bftau) &\te \exp(\bftau\hhat) \\
\bftau =\Log(\cX)    &\te \log(\cX)\vvee 
~. 
\end{align}
See the Appendices for details on the implementation of these maps for different manifolds.

\subsection{Plus and minus operators}

Plus and minus allow us to introduce increments between elements of a (curved) manifold, and express them in its (flat) tangent vector space. 
Denoted by $\op$ and $\om$, they combine one Exp/Log operation with one composition.
Because of the non-commutativity of the composition, they are defined in right- and left- versions depending on the order of the operands.
The right operators are (see \figRef{fig:manifold_q}-\emph{right}), 
\begin{align}
\textrm{right-}\op:&& \cY &= \cX\op{^\cX\!\bftau} \te \cX\circ\Exp({^\cX\!\bftau}) \,\in \cM \label{equ:rplus} \\
\textrm{right-}\om:&& {^\cX\!\bftau} &= \cY\om\cX \,\te \Log(\cX\inv\!\circ\!\cY)\in\mtanat{\cM}{\cX} \label{equ:rminus}
~.
\end{align}
Because in \eqRef{equ:rplus} $\Exp({^\cX\!\bftau})$ appears at the right hand side of the composition, 
${^\cX\!\bftau}$ belongs to the tangent space at $\cX$ (see \eqRef{equ:rminus}): 
we say by convention\footnotemark~that ${^\cX\!\bftau}$ is expressed in the \emph{local} frame at $\cX$ --- we note reference frames with a left superscript.

The left operators are,
\begin{align}
\textrm{left-}\op:&&\cY&={^\cE\!\bftau}\op\cX \te \Exp({^\cE\!\bftau})\circ\cX \in \cM \label{equ:lplus} \\
\textrm{left-}\om:&&{^\cE\!\bftau}&=~\cY\om\cX \te \Log(\cY\!\circ\!\cX\inv)\in\mtanat{\cM}{\cE}
\label{equ:lminus}
~.
\end{align}
Now, in \eqRef{equ:lplus} $\Exp({^\cE\!\bftau})$ is on the left and 
we have ${{^\cE\!\bftau}\in\mtanat{\cM}{\cE}}$: we say that ${{^\cE}\!\bftau}$ is expressed in the \emph{global} frame.

Notice that while left- and right- $\op$ are distinguished by the operands order, the notation $\om$ in \eqRef{equ:rminus} and \eqRef{equ:lminus} is ambiguous.
In this work, we express perturbations locally by default and therefore we use the right- forms of $\op$ and $\om$ by default.
\footnotetext{The convention sticks to that of frame transformation, \eg~${^G\!\bfx=\bfR^L\!\bfx}$, where the matrix $\bfR\in \SO(3)$ transforms local vectors into global.
Notice that this convention is not shared by all authors, and for example \cite{GALLEGO-13} uses the opposite, ${^L\!\bfx=\bfR^G\!\bfx}$.}

\subsection[Adjoint, and adjoint matrix]{The adjoint, and the adjoint matrix}
\label{sec:adjoint}

\begin{figure}[tb]
\centering
\includegraphics{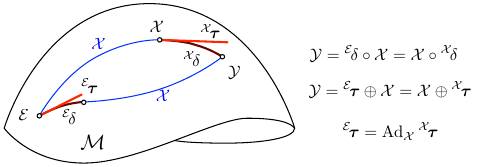}
\caption{Two paths, $\cX\circ{^\cX}\!\delta$ and ${^\cE}\!\delta\circ\cX$, join the origin $\cE$ with the point $\cY$. 
They both compose the element $\cX$ with increments or `deltas' expressed either in the local frame, ${^\cX}\!\delta$, or in the origin, ${^\cE}\!\delta$.
Due to non-commutativity, the elements ${^\cX}\!\delta$ and ${^\cE}\!\delta$ are not equal. 
Their associated tangent vectors ${^\cX}\!\bftau=\Log({^\cX}\!\delta)$ and ${^\cE}\!\bftau=\Log({^\cE}\!\delta)$ are therefore unequal too.
They are related by the linear transform ${^\cE}\!\bftau=\Ad[\cM]{\cX}\,{^\cX}\!\bftau$ where
$\Ad{\cX}$ is the adjoint of $\cM$ at $\cX$. 
}
\label{fig:adjoint}
\end{figure}

If we identify $\cY$ in \eqssRef{equ:rplus,equ:lplus}, we arrive at ${{^\cE\!\bftau}\op\cX} = {\cX\op{^\cX\!\bftau}}$, which determines a relation between the local and global tangent elements (\figRef{fig:adjoint}).
We develop it with \eqssRef{equ:prop_exp,equ:rplus,equ:lplus} as
\begin{align*}
\Exp({^\cE}\!\bftau)\cX 
 &= \cX\Exp({^\cX}\!\bftau) \\
\exp({^\cE}\!\bftau\hhat) 
 &= \cX \exp({^\cX}\!\bftau\hhat) \cX\inv 
  = \exp(\cX {^\cX}\!\bftau\hhat \cX\inv) \\
{^\cE}\!\bftau\hhat 
 &= \cX {^\cX}\!\bftau\hhat \cX\inv 
\end{align*}
\subsubsection{The adjoint}
We thus define the \emph{adjoint} of $\cM$ at $\cX$, noted $\Adh[\cM]{\cX}$, to be
\begin{align}
\Adh{\cX}:\frak{m}\to\frak{m};~~ \bftau\hhat\mapsto \Adh[\cM]{\cX}(\bftau\hhat) \te \cX \bftau\hhat \cX\inv \label{equ:Adjh1} 
~,
\end{align}
so that ${^\cE}\!\bftau\hhat = \Adh[\cM]{\cX}({^\cX}\!\bftau\hhat)$. 
This defines the \emph{adjoint action}  
of the group on its own Lie algebra.
The adjoint has two interesting (and easy to prove) properties,
\begin{align*}
\textrm{Linear} &:& \Adh{\cX}(a\bftau\hhat+b\bfsigma\hhat) =& ~a\Adh{\cX}(\bftau\hhat)\\
&&&+b\Adh{\cX}(\bfsigma\hhat) 
\\
\textrm{Homomorphism} &:& \Adh{\cX}(\Adh{\cY}(\bftau\hhat)) =& ~\Adh{\cX\cY}(\bftau\hhat) 
~.
\end{align*}
\subsubsection{The adjoint matrix}
Since $\Adh{\cX}()$ is linear, we can find an equivalent matrix operator $\Ad{\cX}$ that maps the Cartesian tangent vectors ${^\cE}\!\bftau\cong{^\cE}\!\bftau\hhat$ and ${^\cX}\!\bftau\cong{^\cX}\!\bftau\hhat$,
\begin{align}
\Ad{\cX}:\bbR^m\to\bbR^m;~~ ^\cX\!\bftau\mapsto {^\cE}\!\bftau &= \Ad[\cM]{\cX}{^\cX}\!\bftau \label{equ:Adj2} 
~,
\end{align}
which we call the \emph{adjoint matrix}. This can be computed by applying $\vvee$ to \eqRef{equ:Adjh1}, thus writing
\begin{align}\label{equ:Adj4} 
\Ad[\cM]{\cX}\,\bftau &= (\cX\bftau\hhat\cX\inv)\vvee 
~,
\end{align}
then developing the right hand side to identify the adjoint matrix (see Ex.~\ref{ex:SE3_adjoint} and the appendices).
Additional properties of the adjoint matrix are,
\begin{align}
\cX\op\bftau &= (\Ad[\cM]{\cX}\,\bftau)\op\cX \label{equ:Adj1} \\
\Ad[\cM]{\cX\inv} &= \Ad[\cM]{\cX}\inv \label{equ:Adj5} \\
\Ad[\cM]{\cX\cY} &=\Ad[\cM]{\cX}\Ad[\cM]{\cY} \label{equ:Adj7}
~.
\end{align}
Notice in \eqssRef{equ:Adj5,equ:Adj7} that the left parts of the equality are usually cheaper to compute than the right ones.
We will use the adjoint matrix often as a way to linearly transform vectors of the tangent space at $\cX$ onto vectors of the tangent space at the origin, with ${^\cE}\!\bftau = \Ad{\cX}{^\cX}\!\bftau$,~\eqRef{equ:Adj2}.
In this work, the adjoint matrix will be referred to as simply the adjoint.

\if\examples y

\begin{fexample}{The adjoint matrix of $\SE(3)$}\label{ex:SE3_adjoint}
The $\SE(3)$ group of rigid body motions (see \appRef{sec:SE3}) has group, Lie algebra and vector elements,
\begin{align*}
\bfM&=\begin{bmatrix}\bfR&\bft\\\bf0&1\end{bmatrix}
~,
&
\bftau\hhat&=\begin{bmatrix}\hatx{\bth}&\bfrho\\\bf0&0\end{bmatrix}
~,
&
\bftau &=\begin{bmatrix}\bfrho\\\bth\end{bmatrix}
~.
\end{align*}
The adjoint matrix is identified by developing \eqRef{equ:Adj4} as
\begin{align*}
\Ad{\bfM}\,\bftau
  &= (\bfM\bftau\hhat\bfM\inv)\vvee = \cdots =
  \\
  &= \left(\begin{bmatrix}\bfR\hatx{\bth}\bfR\tr & -\bfR\hatx{\bth}\bfR\tr\bft + \bfR\bfrho \\ \bf0 & \bf0\end{bmatrix}\right)\vvee \\
  &= \left(\begin{bmatrix}\hatx{\bfR\bth} & \hatx{\bft}\bfR\bth + \bfR\bfrho \\ \bf0 & \bf0\end{bmatrix}\right)\vvee \\
  &= \begin{bmatrix}\hatx{\bft}\bfR\bth + \bfR\bfrho \\ \bfR\bth\end{bmatrix} 
  = \begin{bmatrix}\bfR & \hatx{\bft}\bfR\\ \bf0&\bfR\end{bmatrix}\begin{bmatrix}\bfrho \\ \bth\end{bmatrix} 
\end{align*}
where we used $\hatx{\bfR\bth}=\bfR\hatx{\bth}\bfR\tr$ and $\hatx{\bfa}\bfb=-\hatx{\bfb}\bfa$. So the adjoint matrix is 
\begin{align*}
\Ad{\bfM} =  \begin{bmatrix}\bfR & \hatx{\bft}\bfR \\ \bf0&\bfR\end{bmatrix} \quad\in\bbR^{6\times6}
~.
\end{align*}
\end{fexample}		
\fi

\subsection{Derivatives on Lie groups}

Among the different ways to define derivatives in the context of Lie groups, we concentrate on those in the form of Jacobian matrices mapping vector tangent spaces. 
This is sufficient here since in these spaces uncertainties and increments can be properly and easily defined.
Using these Jacobians, the formulas for uncertainty management in Lie groups will largely resemble those in vector spaces.

The Jacobians described hereafter fulfill the chain rule, so that we can easily compute any Jacobian from the partial Jacobian blocks of \emph{inversion}, \emph{composition}, \emph{exponentiation} and \emph{action}.
See \secRef{sec:jacs_chain_rule} for details and proofs.

\subsubsection[Jacobians on vector spaces]{Reminder: Jacobians on vector spaces}

For a multivariate function $f:\bbR^m\to\bbR^n$, the Jacobian matrix is defined as the $n\times m$ matrix stacking all partial derivatives,
\begin{align}\label{equ:jac_Rn_matrix}
\bfJ = \dpar{f(\bfx)}{\bfx} &\te \begin{bmatrix}
\dpar{f_1}{x_1} & \cdots & \dpar{f_1}{x_m} \\
\vdots && \vdots \\
\dpar{f_n}{x_1} & \cdots & \dpar{f_n}{x_m} 
\end{bmatrix} \in\bbR^{n\times m}
~.
\end{align}
It is handy to define this matrix in the following form. Let us partition $\bfJ=[\bfj_1\cdots\bfj_m]$, and let $\bfj_i=[\dpar{f_1}{x_i}\cdots\dpar{f_n}{x_i}]\tr$ be its $i$-th column vector. This column vector responds to
\begin{align}\label{equ:jac_Rn_i}
\bfj_i = \dpar{f(\bfx)}{x_i} \te \lim_{h\to0}\frac{f(\bfx+h\bfe_i)-f(\bfx)}{h} \in \bbR^n
~,
\end{align}
where $\bfe_i$ is the $i$-th vector of the natural basis of $\bbR^m$.
Regarding the numerator, notice that the vector 
\begin{align}\label{equ:jac_Rn_vec}
\bfv_i(h) \te f(\bfx+h\bfe_i)-f(\bfx) \quad \in\bbR^n
\end{align}
is the variation of $f(\bfx)$ when $\bfx$ is perturbed in the direction of $\bfe_i$, and that the respective Jacobian column is just $\bfj_i=\dparil{\bfv_i(h)}{h}|_{h=0}=\lim_{h\to0}\bfv_i(h)/h$.
In this work, for the sake of convenience, we introduce the compact form,
\begin{align}\label{equ:jac_Rn}
\mjac{}{}=\dpar{f(\bfx)}{\bfx}\te\lim_{\bfh\to0}\frac{f(\bfx+\bfh)-f(\bfx)}{\bfh}
\in\bbR^{n\times m}
~,
\end{align}
with $\bfh\in\bbR^m$, which aglutinates all columns \eqRef{equ:jac_Rn_i} 
to form the definition of \eqRef{equ:jac_Rn_matrix}.
We remark that \eqRef{equ:jac_Rn} is just a notation convenience (just as \eqRef{equ:jac_Rn_matrix} is), since division by the vector $\bfh$ is undefined and proper computation requires \eqRef{equ:jac_Rn_i}. 
However, this form may be used to calculate Jacobians by developing the numerator into a form linear in $\bfh$, and identifying the left hand side as the Jacobian, that is,
\begin{align}\label{equ:jac_Rn_identify}
 \lim_{\bfh\to0}\frac{f(\bfx\!+\!\bfh)\!-\!f(\bfx)}{\bfh} 
 = \cdots
 = \lim_{\bfh\to0}\frac{\bfJ\bfh}{\bfh}  
 \te \dpar{\bfJ\bfh}{\bfh} 
 = \bfJ.
\end{align}
Notice finally that for small values of $\bfh$ we have the linear approximation,
\begin{align}
f(\bfx+\bfh) \xrightarrow[\bfh\to0]{} f(\bfx) + \dpar{f(\bfx)}{\bfx}\bfh
~.
\end{align}

\subsubsection{Right Jacobians on Lie goups}

\begin{figure}[tb]
\centering
\includegraphics{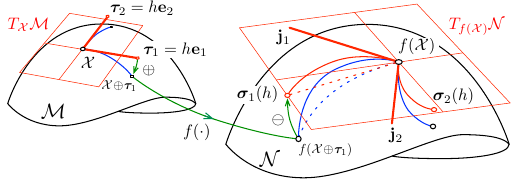}
\caption{Right Jacobian of a function $f:\cM\to\cN$. 
The perturbation vectors in the canonical directions, $\bftau_i=h\bfe_i\in\mtanat{\cM}{\cX}$, are propagated to perturbation vectors $\bfsigma_i\in\mtanat{\cN}{f(\cX)}$ through the processes of plus, apply $f()$, and minus (green arrows), obtaining $\bfsigma_i(h)=f(\cX\op h\bfe_i)\om f(\cX)$. 
For varying values of $h$, notice that in $\cM$ the perturbations $\bftau_i(h)=h\bfe_i$ (thick red) produce paths in $\cM$ (blue) along the geodesic (recall~\figRef{fig:exponential}). 
Notice also that in $\cN$, due to the non-linearity of $f(\cdot)$, the image paths (solid blue) are generally not in the geodesic (dashed blue).
These image paths are lifted onto the tangent space $\mtanat{\cN}{f(\cX)}$, producing smooth curved paths (thin solid red).
The column vectors $\bfj_i$ of $\bfJ$ (thick red) are the derivatives of the lifted paths evaluated at $f(\cX)$, 
\ie, $\bfj_i=\lim_{h\to0}\bfsigma_i(h)/h$.
Each $h\bfe_i\in\mtanat{\cM}{\cX}$ gives place to a $\bfj_i\in\mtanat{\cN}{f(\cX)}$, and thus the resulting Jacobian matrix $\bfJ=[\,\bfj_1\cdots\bfj_m\,]\in\bbR^{n\times m}$ linearly maps vectors from $\mtanat{\cM}{\cX}\cong\bbR^m$ to $\mtanat{\cN}{f(\cX)}\cong\bbR^n$.
}
\label{fig:manifold_g}
\end{figure}

Inspired by the standard derivative definition \eqRef{equ:jac_Rn} above, we can now use our $\op$ and $\om$ operators to define Jacobians of functions $f:\cM\to\cN$ acting on manifolds (see \figRef{fig:manifold_g}).
Using the right- $\{\op,\om\}$ in place of $\{+,-\}$ we obtain a form akin to the standard derivative,\footnote{%
The notation $\ndpar{\cY}{\cX}=\ndpar{f(\cX)}{\cX}$ is chosen in front of other alternatives in order to make the chain rule readable, \ie,~$\ndpar{\cZ}{\cX}=\ndpar{\cZ}{\cY}\ndpar{\cY}{\cX}$.
We will later introduce the lighter notation $\mjac{\cY}{\cX}\te\ndpar{\cY}{\cX}$.
}
\begin{subequations}\label{equ:Jacobian_set}
\begin{align}
\rdpar{f(\cX)}{\cX}
&\te \lim_{\bftau\to0}\frac{f(\cX\op\bftau)\om f(\cX)}{\bftau}
~~~~~ \in\bbR^{n\times m} \label{equ:Jacobian} 
\\
\intertext{which develops as,}
& = \lim_{\bftau\to0}\frac{\Log\big(f(\cX)\inv\circ f(\cX\circ\Exp(\bftau)) \big)}{\bftau} 
\\
&= \pjac{\Log\big(f(\cX)\inv\circ f(\cX\!\circ\!\Exp(\bftau)) \big)}{\bftau}{\bftau=0} 
 \label{equ:Jacobian_aslinear}
.
\end{align}
\end{subequations}
We call this Jacobian the \emph{right Jacobian of $f$}.
Notice that \eqRef{equ:Jacobian_aslinear} is just the standard derivative \eqRef{equ:jac_Rn} of the rather complicated function $g(\bftau)=\Log\big(f(\cX)\inv\circ f(\cX\circ\Exp(\bftau)) \big)$. 
Writing it as in \eqRef{equ:Jacobian} conveys much more intuition: it is the derivative of $f(\cX)$ \wrt $\cX$, only that we expressed the infinitesimal variations in the tangent spaces!
Indeed, thanks to the way right- $\op$ and $\om$ operate, variations in $\cX$ and $f(\cX)$ are now expressed as vectors in the local tangent spaces, \ie, tangent respectively at $\cX\in\cM$ and $f(\cX)\in\cN$. 
This derivative is then a proper Jacobian matrix $\bbR^{n\times m}$ linearly mapping the \emph{local} tangent spaces $\mtanat{\cM}{\cX} \to \mtanat{\cN}{f(\cX)}$ (and we mark the derivative with a local `$\cX$' superscript).
Just as in vector spaces, the columns of this matrix correspond to directional derivatives.
That is, the vector 
\begin{align}
\bfsigma_i(h) &= f(\cX\op h\bfe_i)\om f(\cX) \quad \in\bbR^n \label{equ:jac_N_vec}
\end{align}
(see \figRef{fig:manifold_g} again, and compare $\bfsigma_i$ in \eqRef{equ:jac_N_vec} with $\bfv_i$ in \eqRef{equ:jac_Rn_vec})
is the variation of $f(\cX)$ when $\cX$ varies in the direction of $\bfe_i$.
Its respective Jacobian column is $\bfj_i=\dparil{\bfsigma_i(h)}{h}|_{h=0}$.

As before, we use \eqRef{equ:Jacobian} to actually find Jacobians by resorting to the same mechanism \eqRef{equ:jac_Rn_identify}. 
For example, for a 3D rotation $f:\SO(3)\to\bbR^3;~f(\bfR)=\bfR\bfp$, we have $\cM=\SO(3)$ and $\cN=\bbR^3$ and so (see \appRef{sec:jac_SO3_action}),
\begin{align*}
\rdpar{\bfR\bfp}{\bfR}
 &= \lim_{\bth\to0}\frac{(\bfR\op\bth)\bfp\om\bfR\bfp}{\bth} 
 = \lim_{\bth\to0}\frac{\bfR\Exp(\bth)\bfp-\bfR\bfp}{\bth} \\
 &= \lim_{\bth\to0}\frac{\bfR(\bfI+\hatx{\bth})\bfp-\bfR\bfp}{\bth} 
 = \lim_{\bth\to0}\frac{\bfR\hatx{\bth}\bfp}{\bth} \\
 &= \lim_{\bth\to0}\frac{-\bfR\hatx{\bfp}\bth}{\bth} 
 = -\bfR\hatx{\bfp} 
 ~~~\in \bbR^{3\times 3}
 ~.
\end{align*}
Many examples of this mechanism can be observed in \secRef{sec:derivatives_M} and the appendices.
Remark that whenever the function $f$ passes from one manifold to another, the plus and minus operators in \eqRef{equ:Jacobian} must be selected appropriately: 
$\op$ for the domain $\cM$, and $\om$ for the codomain or image $\cN$.

For small values of $\bftau$, the following approximation holds,
\begin{align}\label{equ:lin_approx}
f(\cX\op{^\cX\!\bftau}) \xrightarrow[{^\cX\!\bftau}\to0]{} f(\cX)\op\rdpar{f(\cX)}{\cX}\,{^\cX\!\bftau}
\quad \in \cN
~.
\end{align}
%

\subsubsection{Left Jacobians on Lie groups}

Derivatives can also be defined from the left- plus and minus operators, leading to,
\begin{align}\label{equ:left-Jacobian}
\ldpar{f(\cX)}{\cX} 
& \te \lim_{\bftau\to0}\frac{f(\bftau\op\cX)\om f(\cX)}{\bftau}  
~~~~~\in\bbR^{n\times m}
\\
& = \lim_{\bftau\to0}\frac{\Log(f(\Exp(\bftau)\circ\cX) \circ f(\cX)\inv)}{\bftau} \notag
\\
&= \pjac{\Log\big(f(\Exp(\bftau)\circ\cX) \circ f(\cX)\inv\big)}{\bftau}{\bftau=0} \notag
~,
\end{align}
which we call the \emph{left Jacobian of $f$}.
Notice that now $\bftau\in\mtanat{\cM}{\cE}$, and the numerator belongs to $\mtanat{\cN}{\cE}$, thus the left Jacobian is a $n\times m$ matrix mapping the \emph{global} tangent spaces, $\mtanat{\cM}{\cE}\to\mtanat{\cN}{\cE}$, which are the Lie algebras of $\cM$ and $\cN$ (and we mark the derivative with a global or origin `$\cE$' superscript).
For small values of $\bftau$ the following holds,
\begin{align}\label{equ:lin_approx_left}
f({^\cE\!\bftau}\op\cX) \xrightarrow[{^\cE\!\bftau}\to0]{} \ldpar{f(\cX)}{\cX}\,{^\cE\!\bftau} \op f(\cX)
\quad \in \cN
~.
\end{align}

\begin{figure}[tb]
\centering
\includegraphics{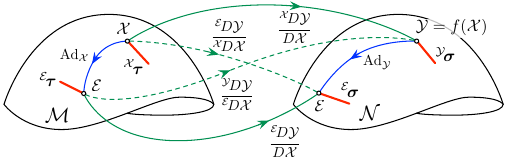}
\caption{Linear maps between all tangent spaces involved in a function $\cY=f(\cX)$, from $\cM$ to $\cN$. The linear maps ${{^\cE}\!\bftau=\Ad[\cM]{\cX}\,{^\cX}\!\bftau}$, ${{^\cE}\!\bfsigma=\Ad[\cN]{\cY}\,{^\cY}\!\bfsigma}$, ${{^\cE}\!\bfsigma=\ldpar{\cY}{\cX}\,{^\cE}\!\bftau}$, and ${{^\cY}\!\bfsigma=\rdpar{\cY}{\cX}\,{^\cX}\!\bftau}$, form a loop (solid) that leads to \eqRef{equ:derivatives_lr_adjoints}. The crossed Jacobians (dashed) form more mapping loops leading to (\ref{equ:derivatives_ex_adjoints},\ref{equ:derivatives_ye_adjoints}).}
\label{fig:jacobians_adjoints}
\end{figure}

We can show from (\ref{equ:Adj1}, \ref{equ:lin_approx}, \ref{equ:lin_approx_left}) (see \figRef{fig:jacobians_adjoints}) that left and right Jacobians are related by the adjoints of $\cM$ and $\cN$,
\begin{align}\label{equ:derivatives_lr_adjoints}
\ldpar{f(\cX)}{\cX}\Ad[\cM]{\cX} = \Ad[\cN]{f(\cX)}\rdpar{f(\cX)}{\cX} 
~.
\end{align}

\subsubsection{Crossed right--left Jacobians}

One can also define Jacobians using right-plus but left-minus, or vice versa. 
Though improbable, these are sometimes useful, since they map local to global tangents or vice versa.
To keep it short, we will just relate them to the other Jacobians through the adjoints,
\begin{align}
\lrdpar{\cY}{\cX} &= \lldpar{\cY}{\cX}\,\Ad{\cX} ~~~\,= \Ad{\cY}\,\rrdpar{\cY}{\cX} \label{equ:derivatives_ex_adjoints}\\
\rldpar{\cY}{\cX} &= \rrdpar{\cY}{\cX}\,\Ad{\cX}\inv = \Ad{\cY}\inv\,\lldpar{\cY}{\cX}\label{equ:derivatives_ye_adjoints}
~,
\end{align}
where $\cY=f(\cX)$. Now, the upper and lower super-scripts indicate the reference frames where the differentials are expressed.
Respective small-tau approximations read,
\begin{align}
f(\cX\op{^\cX}\bftau) 
 &\xrightarrow[{^\cX\!\bftau}\to0]{} \lrdpar{f(\cX)}{\cX}\,{^\cX}\bftau \op f(\cX) \label{equ:lin_approx_rl}\\
f({^\cE}\bftau\op\cX) 
 &\xrightarrow[{^\cE\!\bftau}\to0]{} f(\cX) \op \rldpar{f(\cX)}{\cX}\,{^\cE}\bftau \label{equ:lin_approx_lr}
 ~.
\end{align}

\subsection[Uncertainty, covariances]{Uncertainty in manifolds, covariance propagation}

\begin{figure}[tb]
\centering
\includegraphics{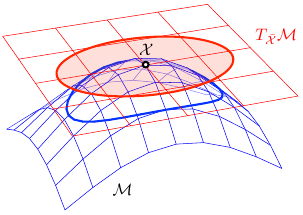}
\caption{Uncertainty around a point $\bar\cX\in\cM$ is properly expressed as a covariance on the vector space tangent at the point (red).
Using $\op$ \eqRef{equ:uncertainty}, the probability ellipses in the tangent space are wrapped over the manifold (blue), thus illustrating the probability concentration region on the group.}
\label{fig:covariance}
\end{figure}

We define local perturbations $\bftau$ around a point $\bar\cX\in\cM$ in the tangent vector space $\mtanat{\cM}{\bar\cX}$, using right- $\op$ and $\om$,
\begin{align}\label{equ:uncertainty}
\cX &= \bar\cX \op \bftau~, & \bftau &=\cX \om \bar\cX ~\in\mtanat{\cM}{\bar\cX}
~.
\end{align}
Covariances matrices can be properly defined on this tangent space at $\bar\cX$ through the standard expectation operator $\bbE[\cdot]$,
\begin{align}\label{equ:cov}
\bfSigma_\cX \te \bbE[\bftau\bftau\tr] = \bbE[(\cX \om \bar\cX)(\cX \om \bar\cX)\tr]~\in\bbR^{m\times m}
~,
\end{align}
allowing us to define Gaussian variables on manifolds, $\cX\sim\cN(\bar\cX,\bfSigma_\cX)$, see \figRef{fig:covariance}.
Notice that although we write $\bfSigma_\cX$, the covariance is rather that of the tangent perturbation $\bftau$.
Since the dimension $m$ of $\mtan{\cM}$ matches the degrees of freedom of $\cM$, these covariances are well defined.%
\footnote{%
A naive definition $\bfSigma_\cX \te \bbE[(\cX - \bar\cX)(\cX - \bar\cX)\tr]$ is always ill-defined if $\mathrm{size}(\cX)>\dim(\cM)$, which is the case for most non-trivial manifolds.%
}

Perturbations can also be expressed in the global reference, that is, in the tangent space at the origin $\mtanat{\cM}{\cE}$, 
using left- $\op$ and $\om$,
\begin{align}\label{equ:uncertainty_left}
\cX &= \bftau\op\bar\cX~, & \bftau &=\cX \om \bar\cX ~\in\mtanat{\cM}{\cE}
~.
\end{align}
This allows global specification of covariance matrices using left-minus in \eqRef{equ:cov}. 
For example, a 3D orientation that is known up to rotations in the horizontal plane can be associated to a covariance $^\cE\bfSigma=\diag(\sigma_\phi^2,\sigma_\theta^2, \infty)$.
Since ``horizontal'' is a global specification, $^\cE\bfSigma$ must be specified in the global reference.

Since global and local perturbations are related by the adjoint \eqRef{equ:Adj2}, their covariances can be transformed with 
\begin{align}
^\cE\bfSigma_{\cX} = \Ad[\cM]{\cX} \, ^\cX\bfSigma_{\cX} \, \Ad[\cM]{\cX}\tr
~.
\end{align}

Covariance propagation through a function $f:\cM\to\cN;\cX\mapsto \cY=f(\cX)$ just requires the linearization \eqRef{equ:lin_approx} with Jacobian matrices \eqRef{equ:Jacobian} to yield the familiar formula,
\begin{align}\label{equ:cov_propagation}
\bfSigma_{\cY} \approx \ndpar{f}{\cX} \, \bfSigma_\cX \, \ndpar{f}{\cX}\tr
~\in\bbR^{n\times n}
~.
\end{align}

\subsection{Discrete integration on manifolds}

\begin{figure}[tb]
\centering
\includegraphics{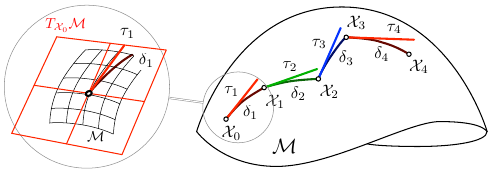}
\caption{Motion integration on a manifold. Each motion data produces a step $\bftau_k\in\mtanat{\cM}{\cX_{k-1}}$, which is wrapped to a local motion increment or `delta' $\delta_k=\Exp(\bftau_k)\in\cM$, and then composed with $\cX_{k-1}$ to yield $\cX_k=\cX_{k-1}\circ\delta_k=\cX_{k-1}\circ\Exp(\bftau_k)=\cX_{k-1}\op\bftau_k\in\cM$.}
\label{fig:manifold_int}
\end{figure}

The exponential map $\cX(t)=\cX_0\circ\Exp(\bfv t)$ performs the continuous-time integral of constant velocities $\bfv\in\mtanat{\cM}{\cX_0}$ onto the manifold.
Non-constant velocities $\bfv(t)$ are typically handled by segmenting them into piecewise constant bits $\bfv_k\in\mtanat{\cM}{\cX_{k-1}}$, of (short) duration $\dt_k$, and writing the discrete integral
\begin{align*}
\cX_k &= \cX_0\circ\Exp(\bfv_1 \dt_1)\circ\Exp(\bfv_1 \dt_2)\circ\cdots\circ\Exp(\bfv_k \dt_k) \\
 &= \cX_0 \op \bfv_1 \dt_1\op\bfv_1 \dt_2\op\cdots\op\bfv_k \dt_k
~.
\end{align*}
Equivalently (\figRef{fig:manifold_int}), we can define $\bftau_k=\bfv_k\dt_k$ and  
construct the integral as a ``sum'' of (small) discrete tangent steps $\bftau_k\in\mtanat{\cM}{\cX_{k-1}}$, \ie,
$
\cX_k \te \cX_0\op\bftau_1\op\bftau_2\op\cdots\op\bftau_k.
$
We write all these variants in recursive form,
\begin{align}\label{equ:int_recursive}
\cX_k = \cX_{k-1}\op\bftau_k = \cX_{k-1}\circ\Exp(\bftau_k) = \cX_{k-1}\circ\Exp(\bfv_k\dt_k)
~.
\end{align}

Common examples are the integration of 3D angular rates $\bfomega$ into the rotation matrix, $\bfR_k=\bfR_{k-1}\Exp(\bfomega_k\dt)$, or into the quaternion, $\bfq_k=\bfq_{k-1}\Exp(\bfomega_k\dt)$.


\section{Differentiation rules on manifolds}
\label{sec:derivatives_M}

For all the typical manifolds $\cM$ that we use, we can determine closed forms for the elementary Jacobians of \emph{inversion}, \emph{composition}, \emph{exponentiation} and \emph{action}.
Moreover, some of these forms can be related to the adjoint $\Ad[\cM]{\cX}$, which becomes a central block of the differentiation process.
Other forms for $\Log$, $\op$ and $\om$ can be easily derived from them.
Once these forms or `blocks' are found, all other Jacobians follow by the chain rule. 
Except for the so-called \emph{left Jacobian}, which we also present below, all Jacobians developed here are right-Jacobians, \ie, defined by \eqRef{equ:Jacobian}.
By following the hints here, the interested reader should find no particular difficulties in developing the left-Jacobians.
For the reader not willing to do this effort, equation \eqRef{equ:derivatives_lr_adjoints} can be used to this end, since
\begin{align}
\ldpar{f(\cX)}{\cX} = \Ad[\cN]{f(\cX)}\rdpar{f(\cX)}{\cX} \Ad[\cM]{\cX}\inv
~.
\end{align}

We use the notations $\mjac{f(\cX)}{\cX}\te\ndpar{f(\cX)}{\cX}$ and $\mjac{\cY}{\cX}\te\ndpar{\cY}{\cX}$.
We notice also that $\Ad[\cM]{\cX}\inv$ should rather be implemented by $\Ad[\cM]{\cX\inv}$ ---see \eqssRef{equ:Adj5,equ:Adj7} and the comment below them.

\subsection{The chain rule}
\label{sec:jacs_chain_rule}

For $\cY=f(\cX)$ and $\cZ=g(\cY)$ we have $\cZ=g(f(\cX))$. 
The chain rule simply states,
\begin{align}
\ndpar{\cZ}{\cX} = \ndpar{\cZ}{\cY}\,\ndpar{\cY}{\cX}
\qquad \textrm{or} \qquad
\mjac{\cZ}{\cX} = \mjac{\cZ}{\cY}\,\mjac{\cY}{\cX}
~.
\end{align}
We prove it here for the right Jacobian using \eqRef{equ:lin_approx} thrice,
\begin{align*}
g(f(\cX))\op\mjac{\cZ}{\cX}\bftau \gets g(f(\cX\op\bftau)) &\to g(f(\cX)~\op\mjac{\cY}{\cX}\bftau) 
 \\
 &\to g(f(\cX))\op\mjac{\cZ}{\cY}\mjac{\cY}{\cX}\bftau
\end{align*}
with the arrows indicating limit as $\bftau\to0$, and so $\mjac{\cZ}{\cX}=\mjac{\cZ}{\cY}\mjac{\cY}{\cX}$. 
The proof for the left and crossed Jacobians is akin, using respectively \eqssRef{equ:lin_approx_left,equ:lin_approx_rl,equ:lin_approx_lr}.
Notice that when mixing right, left and crossed Jacobians, we need to chain also the reference frames, as in \eg\ 
\begin{align}
\rldpar{\cZ}{\cX}
 &= \rrdpar{\cZ}{\cY}\,\rldpar{\cY}{\cX}
 = \rldpar{\cZ}{\cY}\,\lldpar{\cY}{\cX} \label{equ:chain_rule_cross_1}
\\
\lrdpar{\cZ}{\cX}
 &= \lrdpar{\cZ}{\cY}\,\rrdpar{\cY}{\cX}
 = \lldpar{\cZ}{\cY}\,\lrdpar{\cY}{\cX}\label{equ:chain_rule_cross_2}
~,
\end{align}
where the first identity of \eqRef{equ:chain_rule_cross_1} is proven by writing,
\begin{align*}
g(f({^\cE\!\bftau}\op\cX)) 
 &\xrightarrow[{^\cE\!\bftau}\to0]{\eqRef{equ:lin_approx_lr}} 
 g(f(\cX))\op\rldpar{\cZ}{\cX}\,{^\cE}\!\bftau
 ~; \\ 
g(f({^\cE\!\bftau}\op\cX)) 
 &\xrightarrow[{^\cE\!\bftau}\to0]{\eqRef{equ:lin_approx_lr}} 
 g\left(f(\cX)~\op\rldpar{\cY}{\cX}\,{^\cE}\!\bftau\right) \to
 \\
 &\xrightarrow[{^\cE\!\bftau}\to0]{\eqRef{equ:lin_approx}} 
 g(f(\cX))\op\rrdpar{\cZ}{\cY}\,\rldpar{\cY}{\cX}\,{^\cE}\!\bftau 
~,
\end{align*}
and identifying \eqRef{equ:chain_rule_cross_1} in the first and third rows.

\subsection{Elementary Jacobian blocks}
\label{sec:jacs_elementary}

\subsubsection{Inverse}
\label{sec:Jac_inversion}

We define with \eqRef{equ:Jacobian}
\begin{align}
\mjac{\cX\inv}{\cX} 
 &\te \rdpar{\cX\inv}{\cX} \Quad\in \bbR^{m\times m}
 ~. \\
\intertext{This can be determined from the adjoint using  \eqRef{equ:prop_exp} and \eqRef{equ:Adj4},}
\mjac{\cX\inv}{\cX}\label{equ:Jac_inv}
 &= \lim_{\bftau\to0}\frac{\Log((\cX\inv)\inv(\cX\Exp(\bftau))\inv)}{\bftau} \notag\\
 &= \lim_{\bftau\to0}\frac{\Log(\cX\Exp(-\bftau)\cX\inv)}{\bftau} \notag \\
 &= \lim_{\bftau\to0}\frac{(\cX(-\bftau)^\wedge\cX\inv)^\vee}{\bftau} 
 = -\Ad[\cM]{\cX}
~.
\end{align}

\subsubsection{Composition}
\label{sec:Jac_composition}

We define with \eqRef{equ:Jacobian}
\begin{align}
\mjac{\cX\circ\cY}{\cX} &\te \rdpar{\cX\circ\cY}{\cX} &&\in \bbR^{m\times m} \\
\mjac{\cX\circ\cY}{\cY} &\te \rdpar{\cX\circ\cY}{\cY} &&\in \bbR^{m\times m}
~,
\end{align}
and using \eqssRef{equ:prop_exp,equ:Adj4} as above and \eqRef{equ:Adj5},
\begin{align}
\mjac{\cX\circ\cY}{\cX}
&= \lim_{\bftau\to0}\frac{\Log((\cX\cY)\inv(\cX\Exp(\bftau)\cY))}{\bftau} \nonumber\\
&= \lim_{\bftau\to0}\frac{\Log(\cY\inv\Exp(\bftau)\cY)}{\bftau} \nonumber\\
&= \lim_{\bftau\to0}\frac{(\cY\inv\bftau\hhat\cY)\vvee}{\bftau} = \Ad[\cM]{\cY}\inv \label{equ:Jac_comp_1} \\
\mjac{\cX\circ\cY}{\cY} 
&= \cdots = \bfI \label{equ:Jac_comp_2}
\end{align}

\subsubsection[Right and left Jacobians]{Jacobians of $\cM$}

We define the \emph{right Jacobian of $\cM$} as the right Jacobian of $\cX=\Exp(\bftau)$, \ie, for $\bftau\in
\bbR^m$,
\begin{align}
\mjac{}{r}(\bftau) \te \rdpar{\Exp(\bftau)}{\bftau} \in \bbR^{m\times m} 
~,
\label{equ:M_Jr}
\end{align}
which is defined with \eqRef{equ:Jacobian}. 
The right Jacobian maps variations of the argument $\bftau$ into variations in the \emph{local} tangent space at $\Exp(\bftau)$. 
From \eqRef{equ:Jacobian} it is easy to prove that, for small $\delta\bftau$, the following approximations hold,
\begin{align}
\Exp(\bftau+\delta\bftau) &\approx \Exp(\bftau)\Exp(\mjac{}{r}(\bftau)\delta\bftau) \label{equ:Jr_1} \\
\Exp(\bftau)\Exp(\delta\bftau) &\approx \Exp(\bftau+\mjac{-1}{r}(\bftau)\,\delta\bftau) \label{equ:Jr_2} \\
\Log(\Exp(\bftau)\Exp(\delta\bftau)) &\approx \bftau+\mjac{-1}{r}(\bftau)\,\delta\bftau \label{equ:Jr_3}
~.
\end{align}

Complementarily, the \emph{left Jacobian of $\cM$} is defined by, 
\begin{align}
\mjac{}{l}(\bftau) \te \ldpar{\Exp(\bftau)}{\bftau} \in \bbR^{m\times m} 
~,
\label{equ:M_Jl}
\end{align}
using the left Jacobian \eqRef{equ:left-Jacobian}, leading to the approximations
\begin{align}
\Exp(\bftau+\delta\bftau) &\approx \Exp(\mjac{}{l}(\bftau)\delta\bftau)\Exp(\bftau) \label{equ:Jl_1}\\
\Exp(\delta\bftau)\Exp(\bftau) &\approx \Exp(\bftau+\mjac{-1}{l}(\bftau)\,\delta\bftau) \label{equ:Jl_2}\\
\Log(\Exp(\delta\bftau)\Exp(\bftau)) &\approx \bftau+\mjac{-1}{l}(\bftau)\,\delta\bftau \label{equ:Jl_3}
~.
\end{align}
The left Jacobian maps variations of the argument $\bftau$ into variations in the \emph{global} tangent space or Lie algebra. 
From~\eqssRef{equ:Jr_1,equ:Jl_1} we can relate left- and right- Jacobians with the adjoint, 
\begin{align}\label{equ:Jr_Jl_Adj}
\Ad[\cM]{\Exp(\bftau)} = \mjac{}{l}(\bftau)\,\mjac{}{r}\inv(\bftau)
~.
\end{align}
Also, the chain rule allows us to relate $\mjac{}{r}$ and $\mjac{}{l}$, 
\begin{align}\label{equ:Jr_minus}
\mjac{}{r}(-\bftau) 
 &\te \mjac{\Exp(-\bftau)}{-\bftau} 
 = \mjac{\Exp(-\bftau)}{\bftau}\mjac{\bftau}{-\bftau} 
 = \mjac{\Exp(\bftau)\inv}{\bftau}(-\bfI) 
 \notag\\
 &
 = -\mjac{\Exp(\bftau)\inv}{\Exp(\bftau)}\mjac{\Exp(\bftau)}{\bftau} 
 = \Ad{\Exp(\bftau)}\mjac{}{r}(\bftau)
 \notag\\
 &
 = \mjac{}{l}(\bftau)
 ~.
\end{align}

Closed forms of $\mjac{}{r}$, $\mjac{}{r}\inv$, $\mjac{}{l}$ and $\mjac{}{l}\inv$ exist for the typical manifolds in use. 
See the appendices for reference.

\subsubsection{Group action}

For $\cX\in\cM$ and $v\in\cV$, we define with \eqRef{equ:Jacobian} 
\begin{align}
\mjac{\cX\cdot v}{\cX} &\te \rdpar{\cX\cdot v}{\cX} \\
\mjac{\cX\cdot v}{v}  &\te \rdpar{\cX\cdot v}{v} 
~.
\end{align}
Since group actions depend on the set $\cV$, these expressions cannot be generalized. 
See the appendices for reference.

\subsection{Useful, but deduced, Jacobian blocks}

\subsubsection[Log map]{$\Log$ map}

For $\bftau=\Log(\cX)$,
and from \eqRef{equ:Jr_3}, 
\begin{align}\label{equ:Jac_log}
\mjac{\Log(\cX)}{\cX} 
&= \mjac{-1}{r}(\bftau) 
~.
\end{align}

\subsubsection{Plus and minus}

We have
\begin{align}
\mjac{\cX\op\bftau}{\cX}
 &= \mjac{\cX\circ(\Exp(\bftau))}{\cX} 
 ~~~~~~~~~= \Ad[\cM]{\Exp(\bftau)}\inv \\
\mjac{\cX\op\bftau}{\bftau}
 &= \mjac{\cX\circ(\Exp(\bftau))}{\Exp(\bftau)}\mjac{\Exp(\bftau)}{\bftau}
 = \mjac{}{r}(\bftau) 
\end{align}
and given $\cZ=\cX\inv\circ\cY$ and $\bftau=\cY\om\cX=\Log(\cZ)$, 
\begin{align}
\mjac{\cY\om\cX}{\cX}
 &= \mjac{\Log(\cZ)}{\cZ}\mjac{\cZ}{\cX\inv}\mjac{\cX\inv}{\cX} 
  = -\mjac{-1}{l}(\bftau)
 \\
\mjac{\cY\om\cX}{\cY}
 &= \mjac{\Log(\cZ)}{\cZ}\mjac{\cZ}{\cY} 
 ~~~~~~~~~= \mjac{-1}{r}(\bftau)
~.
\end{align}
where the former is proven here
\begin{align*}
\mjac{\cY\om\cX}{\cX}
 &= \mjac{\Log(\cX\inv\circ\cY)}{(\cX\inv\circ\cY)}\,\mjac{(\cX\inv\circ\cY)}{\cX\inv}\,\mjac{\cX\inv}{\cX} 
 \notag\\ 
 (\ref{equ:Jac_log},\ref{equ:Jac_comp_1},\ref{equ:Jac_inv})
 &= ~~\,\mjac{-1}{r}(\bftau)~~\Ad[\cM]{\cY}\inv~~(-\Ad[\cM]{\cX}) 
 \notag\\ 
 (\ref{equ:Adj5},\ref{equ:Adj7})
 &= -\mjac{-1}{r}(\bftau)\,\Ad[\cM]{\cY\inv\cX} 
 \notag\\ 
 &= -\mjac{-1}{r}(\bftau)\,\Ad[\cM]{\Exp(\bftau)}\inv
 \notag\\
 (\ref{equ:Jr_Jl_Adj})
 &
 = -\mjac{-1}{l}(\bftau)
 ~.
\end{align*}


\section{Composite manifolds}

At the price of losing some consistency with the Lie theory, but at the benefit of obtaining some advantages in notation and manipulation, one can consider large and heterogeneous states as manifold composites (or bundles).

\if \examples y 

\begin{fexample}{$\SE(n)$ \emph{vs.} $T(n) \tcross \SO(n)$ \emph{vs.} 
$\langle\bbR^n,\SO(n)\rangle$}
\label{ex:sen_sonxrn_comp}

We consider the space of translations $\bft\in\bbR^n$ and rotations $\bfR\in\SO(n)$. 
We have for this the well-known $\SE(n)$ manifold of rigid motions $\bfM=\begin{bsmallmatrix}\bfR&\bft\\\bf0&1\end{bsmallmatrix}$ (see Apps.~\ref{sec:SE2} and \ref{sec:SE3}), which can also be constructed as $T(n) \tcross \SO(n)$ (see Apps.~\ref{sec:S1_SO2}, \ref{sec:S3_SO3} and \ref{sec:Tn}). 
These two are very similar, but have different tangent parametrizations: 
while $\SE(n)$ uses $\bftau=(\bth,\bfrho)$ with $\bfM=\exp(\bftau^\wedge)$, 
 $T(n) \tcross \SO(n)$ uses $\bftau=(\bth,\bfp)$ with $\bfM=\exp(\bfp^\wedge)\exp(\bth^\wedge)$. 
They share the rotational part $\bth$, but clearly $\bfrho\ne\bfp$ (see \cite[pag.~35]{CHIRIKJIAN-11} for further details).
In short, $\SE(n)$ performs  translation and rotation simultaneously as a continuum, 
while $T(n) \tcross \SO(n)$ performs chained translation+rotation. 
In radical contrast, in the composite $\langle\bbR^n,\SO(n)\rangle$ rotations and translations do not interact at all.
By combining composition with $\Exp()$ we obtain the (right) plus operators,
\begin{align*}
\SE(n)&:& 
\bfM\oplus\bftau &= \begin{bmatrix}
\bfR\Exp(\bth) & \bft+\bfR\bfV(\bth)\bfrho \\
\bf0 & 1
\end{bmatrix} 
\\
T(n) \tcross \SO(n)&:&
\bfM\oplus\bftau &= \begin{bmatrix}
\bfR\Exp(\bth) & \bft+\bfR\bfp \\
\bf0 & 1
\end{bmatrix} 
\\ 
\langle\bbR^n,\SO(n)\rangle&:&
\bfM\dplus\bftau 
& 
= 
\begin{bmatrix}
\bft+\bfp \\
\bfR\Exp(\bth)
\end{bmatrix} 
\end{align*}
where either $\oplus$ may be used for the system dynamics, \eg~motion integration, but usually not $\dplus$, which might however be used to model perturbations.
Their respective minus operators read,
\begin{align*}
\SE(n)&:&
\bfM_2\ominus\bfM_1 & = \begin{bmatrix}
\bfV_1\inv\bfR_1\tr(\bfp_2 - \bfp_1) \\ \Log(\bfR_1\tr\bfR_2)
\end{bmatrix} 
\\
T(n) \tcross \SO(n)&:&
\bfM_2\ominus\bfM_1 & = \begin{bmatrix}
\bfR_1\tr(\bfp_2 - \bfp_1) \\ \Log(\bfR_1\tr\bfR_2)
\end{bmatrix} 
\\
\langle\bbR^n,\SO(n)\rangle&:&
\bfM_2\dminus\bfM_1 
&= 
\begin{bmatrix}
\bfp_2 - \bfp_1 \\ \Log(\bfR_1\tr\bfR_2)
\end{bmatrix} 
~,
\end{align*}
where now, interestingly, $\dminus$ can be used to evaluate errors and uncertainty. This makes $\dplus,\dminus$ valuable operators for computing derivatives and covariances.
\end{fexample}
 \fi

A \emph{composite manifold} $\cM=\langle\cM_1,\cdots,\cM_M\rangle$ is no less than the concatenation of $M$ non-interacting manifolds.
This stems from defining identity, inverse and composition acting on each block of the composite separately,
\begin{align}
\cE_\diamond &\te \begin{bmatrix}
\cE_1 \\ \vdots \\ \cE_M
\end{bmatrix},
&
\cX^\diamond &\te \begin{bmatrix}
\cX\inv \\ \vdots \\ \cX_M\inv
\end{bmatrix},
&
\cX\diamond\cY &\te \begin{bmatrix}
\cX\circ\cY_1 \\
\vdots\\
\cX_M\circ\cY_M 
\end{bmatrix}
,
\end{align}
thereby fulfilling the group axioms, as well as a non-interacting retraction map, which we will also note as ``exponential map" for the sake of unifying notations (notice the angled brackets),
\begin{align}\label{equ:exp_composite}
\Exp\langle\bftau\rangle &\te \begin{bmatrix}
\Exp(\bftau_1) \\ \vdots \\ \Exp(\bftau_M)
\end{bmatrix}
\,,
&
\Log\langle\cX\rangle &\te \begin{bmatrix}
\Log(\cX) \\ \vdots \\ \Log(\cX_M)
\end{bmatrix}
,
\end{align}
thereby ensuring smoothness.
These yield the composite's right- plus and minus (notice the diamond symbols),
\begin{align}
\cX\dplus\bftau &\te \cX\diamond\Exp\langle\bftau\rangle \\
\cY\dminus\cX &\te \Log\langle\cX^\diamond\diamond\cY\rangle
~.
\end{align}

The key consequence of these considerations%
\if\examples y{ (see Ex.~\ref{ex:sen_sonxrn_comp}) }\else { }\fi 
is that new derivatives can be defined,\footnotemark\ using $\dplus$ and $\dminus$,
\footnotetext{We assume here right derivatives, but the same applies to left derivatives.}
\begin{align}
\ndpar{f(\cX)}{\cX} \te \lim_{\bftau\to0}\frac{f(\cX\dplus\bftau)\dminus f(\cX)}{\bftau}
\label{equ:Jacobian_composite}
~.
\end{align}
With this derivative, Jacobians of functions $f:\cM\to\cN$ acting on composite manifolds can be determined in a per-block basis, which yields simple expressions requiring only knowledge on the manifold blocks of the composite,
\begin{align}
\ndpar{f(\cX)}{\cX} &= \begin{bmatrix}
\ndpar{f_1}{\cX_1} & \cdots & \ndpar{f_1}{\cX_M} \\
\vdots             & \ddots & \vdots \\
\ndpar{f_N}{\cX_1} & \cdots & \ndpar{f_N}{\cX_M} \\
\end{bmatrix}
~,
\end{align}
where $\ndpar{f_i}{\cX_j}$ are each computed with \eqRef{equ:Jacobian}. 
For small values of $\bftau$ the following holds,
\begin{align}\label{equ:lin_approx_composite}
f(\cX\dplus\bftau) \xrightarrow[{\bftau}\to0]{} f(\cX)\dplus\ndpar{f(\cX)}{\cX}\,{\bftau}
\quad \in \cN
~.
\end{align}

When using these derivatives, covariances and uncertainty propagation must follow the convention. In particular, the covariance matrix \eqRef{equ:cov} becomes
\begin{align}\label{equ:cov_composite}
\bfSigma_\cX \te \bbE[(\cX \dminus \bar\cX)(\cX \dminus \bar\cX)\tr]~\in\bbR^{n\times n}
~,
\end{align}
for which the linearized propagation \eqRef{equ:cov_propagation} using \eqRef{equ:Jacobian_composite} applies.


\newcommand{\dx}{{\delta\bfx}}

\section{Landmark-based localization and mapping}
\label{sec:SLAM}

We provide three applicative examples of the theory for robot localization and mapping. 
The first one is a Kalman filter for landmark-based localization.
The second one is a graph-based smoothing method for simultaneous localization and mapping.
The third one adds sensor self-calibration.
They are based on a common setup, explained as follows.

We consider a robot in the plane
(see \secRef{sec:demos_3D} for the 3D case)
surrounded by a small number of punctual landmarks or \emph{beacons}. 
The robot receives control actions in the form of axial and angular velocities and is able to measure the location of the beacons \wrt its own reference frame.

The robot pose is in $\SE(2)$ (\appRef{sec:SE2}) and the beacon positions in $\bbR^2$ (\appRef{sec:Tn}),
\begin{align*}
\cX &= 
\begin{bmatrix}
\bfR & \bft \\ \bf0 & 1
\end{bmatrix}\in\SE(2)
~,
&
\bfb_k &= \begin{bmatrix}
x_k \\ y_k
\end{bmatrix}\in\bbR^2
~.
\end{align*}

The control signal $\bfu$ is a twist in $\se(2)$ comprising longitudinal velocity $v$ and angular velocity $\omega$, with no lateral velocity component, integrated over the sampling time $\dt$. 
The control is corrupted by additive Gaussian noise $\bfw\sim\cN(\bf0,\bfW)$.
This noise accounts for possible lateral wheel slippages $u_s$ through a value of $\sigma_s\ne0$, 
\begin{align}
\bfu &
= \begin{bmatrix} u_v \\ u_s \\ u_\omega \end{bmatrix} 
= \begin{bmatrix} v\,\dt \\ 0 \\ \omega\,\dt \end{bmatrix}+\bfw  
&&\in\se(2)
\\
\bfW &= \begin{bmatrix}
\sigma_v^2\dt &0&0 \\ 0&\sigma_s^2\dt&0 \\ 0&0&\sigma_w^2\dt
\end{bmatrix} &&\in\bbR^{3\times3}.
\end{align}
At the arrival of a control $\bfu_j$ at time $j$, the robot pose is updated with \eqRef{equ:int_recursive},
\begin{align}\label{equ:motion}
\cX_j &= \cX_i \op \bfu_j \te \cX_i \Exp(\bfu_j)
~.
\end{align}

Landmark measurements are of the range and bearing type, though they are put in Cartesian form for simplicity. 
Their noise $\bfn\sim\cN({\bf0},\bfN)$ is zero mean Gaussian,
\begin{align}\label{equ:meas_beacon}
\bfy_k &= \cX\inv\cdot\bfb_k + \bfn = \bfR\tr(\bfb_k-\bft) + \bfn	&&\in \bbR^2
\\
\bfN &= \begin{bmatrix}
\sigma_x^2 &0 \\ 0& \sigma_y^2
\end{bmatrix}										&&\in\bbR^{2\times2}
~,
\end{align}
where we notice the rigid motion action $\cX\inv\cdot\bfb_k$ (see \appRef{sec:SE2}).

\subsection{Localization with error-state Kalman filter on manifold}
\label{sec:loc_ESKF}

We initially consider the beacons $\bfb_k$ situated at known positions. 
We define the pose to estimate as $\hat\cX\in\SE(2)$. 
The estimation error $\dx$ and its covariance $\bfP$ are expressed in the tangent space at $\hat\cX$ with \eqssRef{equ:uncertainty,equ:cov}, 
\begin{align}\label{equ:loc_Gaussian}
\dx &\te \cX\ominus\hat\cX && \in\bbR^3
\\
\bfP &\te \bbE[(\cX\ominus\hat\cX)(\cX\ominus\hat\cX)\tr]&& \in\bbR^{3\times3}
~.
\end{align}
At each robot motion we apply ESKF prediction,
\begin{align}
\hat\cX_j &= \hat\cX_i\op\bfu_j \label{equ:loc_ekf_pred}
\\
\bfP_j &= \bfF\,\bfP_i\,\bfF\tr + \bfG\,\bfW_j\,\bfG\tr
~,
\end{align}
with the Jacobians 
computed from the blocks in \appRef{sec:SE2}, 
\begin{align*}
\bfF &\te \mjac{\cX_j}{\cX_i} = \mjac{\hat\cX_i\op\bfu_j}{\hat\cX_i} 
= \Ad{\Exp(\bfu_j)}\inv
\\
\bfG &\te \mjac{\cX_j}{\bfu_j} = \mjac{\hat\cX_i\op\bfu_j}{\bfu_j} = \mjac{}{r}(\bfu_j)
~.
\end{align*}
At each beacon measurement $\bfy_k$ we apply ESKF correction,
\begin{align}
\textrm{Innovation}&:&\bfz &= \bfy_k - \hat\cX\inv\cdot\bfb_k \notag\\
\textrm{Innovation cov.}&:&\bfZ &= \bfH\,\bfP\,\bfH\tr+\bfN \notag\\
\textrm{Kalman gain}&:&\bfK &= \bfP\,\bfH\tr\,\bfZ\inv \notag\\
\textrm{Observed error}&:&\dx &= \bfK\bfz \notag\\
\textrm{State update}&:&\hat\cX &\gets \hat\cX \oplus \dx \label{equ:loc_ekf_corr}\\
\textrm{Cov. update}&:&\bfP &\gets \bfP - \bfK\,\bfZ\,\bfK\tr
~,
\end{align}
with the Jacobian computed from the blocks in \appRef{sec:SE2}, 
\begin{align*}
\bfH 
&\te \mjac{\cX\inv\cdot\bfb_k}{\cX}= \mjac{\cX\inv\cdot\bfb_k}{\cX\inv}\,\mjac{\cX\inv}{\cX} 
\notag\\
&= 
\begin{bmatrix}\bfR\tr & \bfR\tr\hatx{1}\bfb_k\end{bmatrix}
\begin{bmatrix}
-\bfR & \hatx{1}\bft \\ \bf0 & -1
\end{bmatrix}
\notag\\
&=
-\begin{bmatrix}\bfI & \bfR\tr\hatx{1}(\bfb_k-\bft)\end{bmatrix}
~.
\end{align*}

Notice that the only changes \wrt a regular EKF are in \eqRef{equ:loc_ekf_pred} and \eqRef{equ:loc_ekf_corr}, where regular $+$ are substituted by $\op$. 
The Jacobians on the contrary are all computed using the Lie theory (see \appRef{sec:SE2}).
Interstingly, their usage is the same as in standard EKF --- see \eg\ the equation of the Kalman gain, which is the standard $\bfK = \bfP\bfH\tr(\bfH\bfP\bfH\tr+\bfN)\inv$.

\subsection{Smooting and Mapping with graph-based optimization}
\label{sec:SAM}

\begin{figure}
\centering
\includegraphics{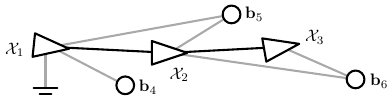}
\caption{SAM factor graph with 3 poses and 3 beacons. 
Each measurement contributes a factor in the graph. 
There are 2 motion factors (black) and 5 beacon factors (gray). 
A prior factor on $\cX_1$ provides global observability.}
\label{fig:graph-SLAM}
\end{figure}

We consider now the problem of smoothing and mapping (SAM), where the variables to estimate are the beacons' locations and the robot's trajectory.
The solver of choice is a graph-based iterative least-squares optimizer. 
For simplicity, we assume the trajectory comprised of three robot poses $\{\cX_1\cdots\cX_3\}$, and a world with three beacons $\{\bfb_4\cdots\bfb_6\}$. 
The problem state is the composite
\begin{align}\label{equ:SAM_state}
\cX &= \langle
\cX_1 , \cX_2 , \cX_3 , \bfb_4 , \bfb_5 , \bfb_6
\rangle, \quad \cX_i\in\SE(2),\quad\bfb_k\in\bbR^2.
\end{align}
The resulting factor graph \cite{DELLAERT-IJRR-06} is shown in \figRef{fig:graph-SLAM}. 
Each prior or measurement contributes a factor in the graph.
Motion measurements from pose $i$ to $j$ are derived from \eqRef{equ:motion},
while measurements of beacon $k$ from pose $i$ respond to \eqRef{equ:meas_beacon},
\begin{align}
\bfu_{ij} &= \cX_j\ominus\cX_i + \bfw_{ij} = \Log(\cX_i\inv\cX_j) + \bfw_{ij} \label{equ:meas_motion} \\
\bfy_{ik} &= \cX_i\inv\cdot\bfb_k + \bfn_{ik}
~.
\end{align}
Each factor comes with an information matrix, $\bfOmega_{1} \te \bfW_{1}\inv$, $\bfOmega_{ij} \te \bfW_{ij}\inv$ and $\bfOmega_{ik} \te \bfN_{ik}\inv$.
The expectation residuals are, 
\begin{align}
\textrm{prior residual}&:&\bfr_{1}(\cX) &= \bfOmega_{1}^{\top/2}(\cX_1 \om \hat\cX_1)
\notag
\\
\textrm{motion residual}&:&\bfr_{ij}(\cX) &= \bfOmega_{ij}^{\top/2}(\bfu_{ij} - (\hat\cX_j\ominus\hat\cX_i))
\notag
\\
\textrm{beacon residual}&:&\bfr_{ik}(\cX) &= \bfOmega_{ik}^{\top/2}(\bfy_{ik} - \hat\cX_i\inv\cdot\hat\bfb_k)
\notag
~.
\end{align}
The optimum update step $\dx$ stems from minimizing
\begin{align}\label{equ:SAM_problem}
\dx^* = \argmin_{\dx} \sum_{p\in\cP} \bfr_p(\cX\dplus\dx)\tr\bfr_p(\cX\dplus\dx)
\end{align}
with $\cP=\{1,12,23,14,15,25,26,36\}$ the set of node pairs of each measurement (see \figRef{fig:graph-SLAM}).
The problem is solved iteratively as follows. 
Each residual in the sum \eqRef{equ:SAM_problem} is linearized to $\bfr_p(\cX\dplus\dx)\approx\bfr_p(\cX)\dplus\mjac{\bfr_p}{\cX}\dx$ following \eqRef{equ:lin_approx_composite}, where $\mjac{\bfr_p}{\cX}$ are sparse Jacobians.
The non-zero blocks of these Jacobians, that is $\mjac{\bfr_{1}}{\cX_1}$, $\mjac{\bfr_{ij}}{\cX_i}$, $\mjac{\bfr_{ij}}{\cX_j}$, $\mjac{\bfr_{ik}}{\cX_i}$ and $\mjac{\bfr_{ik}}{\bfb_k}$, can be easily computed following the methods in \secRef{sec:loc_ESKF}, and noticing that by definition $\mjac{f(\cX\op\dx)}{\dx}|_{\dx=0}=\mjac{f(\cX\op\dx)}{\cX}|_{\dx=0}=\mjac{f(\cX)}{\cX}$.
Building the total Jacobian matrix and residual vector,
\begin{align}\label{equ:SAM_problem_lin}
\bfJ &= \begin{bmatrix}
\mjac{\bfr_{1}}{\cX_1} & \bf0 & \bf0 & \bf0 & \bf0 & \bf0 \\ 
\mjac{\bfr_{12}}{\cX_1} & \mjac{\bfr_{12}}{\cX_2} & \bf0 & \bf0 & \bf0 & \bf0 \\ 
\bf0 & \mjac{\bfr_{23}}{\cX_2} & \mjac{\bfr_{23}}{\cX_3} & \bf0 & \bf0 & \bf0 \\ 
\mjac{\bfr_{14}}{\cX_1} & \bf0 & \bf0 & \mjac{\bfr_{14}}{\bfb_4} & \bf0 & \bf0 \\ 
\mjac{\bfr_{15}}{\cX_1} & \bf0 & \bf0 & \bf0 & \mjac{\bfr_{15}}{\bfb_5} & \bf0 \\ 
\bf0 & \mjac{\bfr_{25}}{\cX_2} & \bf0 & \bf0 & \mjac{\bfr_{25}}{\bfb_5} & \bf0 \\ 
\bf0 & \mjac{\bfr_{26}}{\cX_2} & \bf0 & \bf0 & \bf0 & \mjac{\bfr_{26}}{\bfb_6} \\ 
\bf0 & \bf0 & \mjac{\bfr_{36}}{\cX_3} & \bf0 & \bf0 & \mjac{\bfr_{36}}{\bfb_6} 
\end{bmatrix}
&
\bfr &= \begin{bmatrix}
\bfr_{1} \\
\bfr_{12} \\
\bfr_{23} \\
\bfr_{14} \\
\bfr_{15} \\
\bfr_{25} \\
\bfr_{26} \\
\bfr_{36}
\end{bmatrix}
\end{align}
the linearized \eqRef{equ:SAM_problem} is now transformed \cite{DELLAERT-IJRR-06} to minimizing  
\begin{align}
\dx^* 
 &= \argmin_\dx \norm{\bfr+\bfJ\dx}^2
.
\end{align}
This is solved via least-squares using the pseudoinverse of $\bfJ$ (for large problems, QR \cite{DELLAERT-IJRR-06,KAESS-11-ISAM2} or Cholesky \cite{KUMMERLE-11-G2O,ILA-17_SLAM++} factorizations are required), 
\begin{align}
\dx^* &= -(\bfJ\tr\bfJ)\inv\bfJ\tr\bfr \label{equ:SAM_opt_step}
~,
\intertext{yielding the optimal step $\dx^*$ used to update the state,}
\cX &\gets \cX \dplus \dx^* \label{equ:SAM_update}
~.
\end{align}
The procedure is iterated until convergence.

We highlight here the use of the composite notation in \eqRef{equ:SAM_state}, which allows block-wise definitions of the Jacobian \eqRef{equ:SAM_problem_lin} and the update \eqRef{equ:SAM_update}. We also remark the use of the $\SE(2)$ manifold in the motion and measurement models, as we did in the ESKF case in \secRef{sec:loc_ESKF}.

\subsection{Smoothing and mapping with self-calibration}

We consider the same problem as above but with a motion sensor affected by an unknown calibration bias $\bfc=(c_v,c_\omega)\tr$,
 so that the control is now $
\tilde\bfu 
=
(v\dt + c_v ,~
0 ,~
\omega \dt + c_\omega)\tr + \bfw
$.
We define the bias correction function $c()$,
\begin{align}\label{equ:bias}
\bfu &= c\,(\tilde\bfu, \bfc) \te \begin{bmatrix}
\tilde u_v - c_v \\
\tilde u_s \\
\tilde u_\omega - c_\omega
\end{bmatrix} \quad \in \bbR^3\cong\se(2)
~.
\end{align}
The state composite is augmented with the unknowns $\bfc$,
\begin{align*}
\cX &= \langle
\bfc, \cX_1 , \cX_2 , \cX_3 , \bfb_4 , \bfb_5 , \bfb_6
\rangle
~, 
\\
\bfc&\in\bbR^2,\qquad\cX_i\in\SE(2),\qquad\bfb_k\in\bbR^2
~,
\end{align*}
and the motion residual becomes
\begin{align*}
\bfr_{ij}(\cX) &= \bfOmega_{ij}^{\top/2}\big(c\,(\tilde\bfu_{ij} , \bfc) - (\hat\cX_j\ominus\hat\cX_i)\big)
~.
\end{align*}
The procedure is as in \secRef{sec:SAM} above, and just the total Jacobian is modified with an extra column on the left,
\begin{align*}
\bfJ &= \begin{bmatrix}
\bf0 & \mjac{\bfr_{1}}{\cX_1} & \bf0 & \bf0 & \bf0 & \bf0 & \bf0 \\ 
\mjac{\bfr_{12}}{\bfc} & \mjac{\bfr_{12}}{\cX_1} & \mjac{\bfr_{12}}{\cX_2} & \bf0 & \bf0 & \bf0 & \bf0 \\ 
\mjac{\bfr_{23}}{\bfc} & \bf0 & \mjac{\bfr_{23}}{\cX_2} & \mjac{\bfr_{23}}{\cX_3} & \bf0 & \bf0 & \bf0 \\ 
\bf0 & \mjac{\bfr_{14}}{\cX_1} & \bf0 & \bf0 & \mjac{\bfr_{14}}{\bfb_4} & \bf0 & \bf0 \\ 
\bf0 & \mjac{\bfr_{15}}{\cX_1} & \bf0 & \bf0 & \bf0 & \mjac{\bfr_{15}}{\bfb_5} & \bf0 \\ 
\bf0 & \bf0 & \mjac{\bfr_{25}}{\cX_2} & \bf0 & \bf0 & \mjac{\bfr_{25}}{\bfb_5} & \bf0 \\ 
\bf0 & \bf0 & \mjac{\bfr_{26}}{\cX_2} & \bf0 & \bf0 & \bf0 & \mjac{\bfr_{26}}{\bfb_6} \\ 
\bf0 & \bf0 & \bf0 & \mjac{\bfr_{36}}{\cX_3} & \bf0 & \bf0 & \mjac{\bfr_{36}}{\bfb_6} 
\end{bmatrix}
~,
\end{align*}
where $\mjac{\bfr_{ij}}{\bfc} = \bfOmega_{ij}^{\top/2}\mjac{c(\bfu_{ij} , \bfc)}{\bfc}$, with $\mjac{c(\bfu_{ij} , \bfc)}{\bfc}$ the $3\times 2$ Jacobian of \eqRef{equ:bias}. 
The optimal solution is obtained with \eqssRef{equ:SAM_opt_step,equ:SAM_update}.
The resulting optimal state $\cX$ includes an optimal estimate of $\bfc$, that is, the self-calibration of the sensor bias.

\subsection{3D implementations}
\label{sec:demos_3D}

It is surprisingly easy to bring all the examples above to 3D. 
It suffices to define all variables in the correct spaces:
$\cX\in\SE(3)$ and $\bfu\in\bbR^6\cong\se(3)$ (\appRef{sec:SE3}), and $\{\bfb_k,\bfy\}\in\bbR^3$ (\appRef{sec:Tn}).
Jacobians and covariances matrices will follow with appropriate sizes.
The interest here is in realizing that all the math in the algorithms, that is from \eqRef{equ:loc_Gaussian} onwards, is exactly the same for 2D and 3D: the abstraction level provided by the Lie theory has made this possible.



\section{Conclusion}

We have presented the essential of Lie theory in a form that should be useful for an audience skilled in state estimation, with a focus on robotics applications.
This we have done through several initiatives: 

First, a selection of materials that avoids abstract mathematical concepts as much as possible. This helps to focus Lie theory to make its tools easier to understand and to use.

Second, we chose a didactical approach, with significant redundancy. The main text is generic and covers the abstract points of Lie theory. It is accompanied by boxed examples, which ground the abstract concepts to particular Lie groups, and plenty of figures with very verbose captions.

Third, we have promoted the usage of handy operators, such as the capitalized $\Exp()$ and $\Log()$ maps, and the plus and minus operators $\op,\,\om, \dplus, \dminus$. They allow us to work on the Cartesian representation of the tangent spaces, producing formulas for derivatives and covariance handling that greatly resemble their counterparts in standard vector spaces. 

Fourth, we have made special emphasis on the definition, geometrical interpretation, and computation of Jacobians. For this, we have introduced notations for the Jacobian matrices and covariances that allow a manipulation that is visually powerful. In particular, the chain rule is clearly visible with this notation. This helps to build intuition and reducing errors.

Fifth, we present in the appendices that follow an extensive compendium of formulas for the most common groups in robotics. In 2D, we present the rotation groups of unit complex numbers $S^1$ and rotation matrices $\SO(2)$, and the rigid motion group $\SE(2)$. In 3D, we present the groups of unit quaternions $S^3$ and rotation matrices $\SO(3)$, both used for rotations, and the rigid motion group $\SE(3)$. We also present the translation groups for any dimension, which can be implemented by either the standard vector space $\bbR^n$ under addition, or by the matrix translation group $T(n)$ under multiplication.

Sixth, we have presented some applicative examples to illustrate the capacity of Lie theory to solve robotics problems with elegance and precision. 
The somewhat naive concept of composite group helps to unify heterogeneous state vectors into a Lie-theoretic form. 

Finally, we accompany this text with the new C++ library \manif\ \cite{DERAY-20-manif} implementing the tools described here. \manif\ can be found at \url{https://github.com/artivis/manif}.
The applications in \secRef{sec:SLAM} are demonstrated in \manif\ as examples.

Though we do not introduce any new theoretical material, we believe the form in which Lie theory is here exposed will help many researchers enter the field for their future developments.
We also believe this alone represents a valuable contribution.


\begin{appendices}


\section{The 2D rotation groups $S^1$ and $SO(2)$}
\label{sec:S1_SO2}

The Lie group $S^1$ is the group of unit complex numbers under the complex product. 
Its topology is the unit circle, or the unit 1-sphere, and therefore the name $S^1$. 
The group, Lie algebra and vector elements have the form,
\begin{align}
\bfz&=\cos\theta+i\sin\theta, & \tau^\wedge&=i\theta, & \tau&=\theta
~.
\end{align}
Inversion and composition are achieved by conjugation $\bfz\inv = \bfz^*$, and product $\bfz_a\circ\bfz_b = \bfz_a\,\bfz_b$.

The group $\SO(2)$ is the group of special orthogonal matrices in the plane, or rotation matrices, under matrix multiplication.
Group, Lie algebra and vector elements have the form,
\begin{align}
\bfR&= \begin{bsmallmatrix}
 \cos\theta & -\sin\theta \\ \sin\theta & \cos\theta 
 \end{bsmallmatrix}
, & \tau^\wedge&=\hatx{\theta}\te \begin{bsmallmatrix}
0 & -\theta \\ \theta & 0
\end{bsmallmatrix}, & \tau&=\theta
~.
\end{align}
Inversion and composition are achieved by transposition $\bfR\inv = \bfR\tr$, and product $\bfR_a\circ\bfR_b = \bfR_a\,\bfR_b$.

Both groups rotate 2-vectors, and they have isomorphic tangent spaces.
We thus study them together.

\subsection{Exp and Log maps}

Exp and Log maps may be defined for complex numbers of $S^1$ and rotation matrices of $SO(2)$. 
For $S^1$ we have,
\begin{align}
\bfz = \Exp(\theta) &= \cos\theta+i\sin\theta && \in\bbC \label{equ:Euler_formula}\\
\theta = \Log(\bfz) &= \arctan(\Im(\bfz),\Re(\bfz)) && \in\bbR
~,
\intertext{where \eqRef{equ:Euler_formula} is the Euler formula, whereas for $SO(2)$,}
\bfR = \Exp(\theta) &= \begin{bmatrix}
\cos\theta & -\sin\theta \\ \sin\theta & \cos\theta
\end{bmatrix} &&\in\bbR^{2\tcross2} \label{equ:R_SO2} \\
\theta = \Log(\bfR) &= \arctan(r_{21},r_{11}) && \in\bbR
~.
\end{align}

\subsection{Inverse, composition, exponential map}

We consider generic 2D rotation elements, and note them with the sans-serif font, $\sQ,\sR$. 
We have
\begin{align}
\sR(\theta)\inv &= \sR(-\theta) \\
\sQ\circ\sR   &= \sR\circ\sQ 
~,
\intertext{\ie, planar rotations are commutative. 
It follows that}
\Exp(\theta_1+\theta_2) &= \Exp(\theta_1)\circ\Exp(\theta_2) \\
\Log(\sQ\circ\sR) &= \Log(\sQ)+\Log(\sR) \\
\sQ\om\sR &= \theta_Q-\theta_R 
~.
\end{align}

\subsection{Jacobian blocks}
\label{sec:derivatives_SO2}

Since our defined derivatives map tangent vector spaces, and these spaces coincide for the planar rotation manifolds of $S^1$ and $SO(2)$, \ie, $\theta=\Log(\bfz)=\Log(\bfR)$, it follows that the Jacobians are independent of the representation used ($\bfz$ or $\bfR$). 

\subsubsection[Adjoint and other Jacobians]{Adjoint and other trivial Jacobians}\label{sec:SO2_jacs}
From \eqRef{equ:Jacobian}, \secRef{sec:jacs_elementary} and the properties above, the following scalar derivative blocks become trivial,
\begin{align}
\Ad[\SO(2)]{\sR} &= 1 && \in\bbR \\
\mjac{\sR\inv}{\sR} 
 &= -1 && \in \bbR\\
\mjac{\sQ\circ\sR}{\sQ} 
 = \mjac{\sQ\circ\sR}{\sR} 
 &= 1 && \in \bbR\\
\bfJ_r(\theta)
 = \bfJ_l(\theta)
 &= 1 && \in \bbR \\
\mjac{\sR\op\theta}{\sR}   
 =~~\,\mjac{\sR\op\theta}{\theta}   
 & =1 && \in \bbR\\
\mjac{\sQ\om\sR}{\sQ} 
 = -\mjac{\sQ\om\sR}{\sR} 
 &= 1 && \in \bbR
\end{align}
%

\subsubsection{Rotation action}
\label{sec:jac_SO2_action}

For the action $\sR\cdot\bfv$ we have,
\begin{align}
\mjac{\sR\cdot\bfv}{\sR}
&= \lim_{\theta\to0}\frac{\bfR\Exp(\theta)\bfv-\bfR\bfv}{\theta} \notag \\
&= \lim_{\theta\to0}\frac{\bfR(\bfI+\hatx{\theta})\bfv-\bfR\bfv}{\theta} \notag \\
&= \lim_{\theta\to0}\frac{\bfR\hatx{\theta}\bfv}{\theta} 
 = \bfR\hatx{1}\bfv && \in \bbR^{2\times 1} 
\intertext{and}
\mjac{\sR\cdot\bfv}{\bfv} &= \ndpar{\bfR\bfv}{\bfv} = \bfR && \in\bbR^{2\times2}
~.
\end{align}
%


\section{The 3D rotation groups $S^3$ and $SO(3)$}
\label{sec:S3_SO3}

The Lie group $S^3$ is the group of unit quaternions under quaternion multiplication. 
Its topology is the unit 3-sphere in $\bbR^4$, and therefore its name $S^3$.
Quaternions (please consult \cite{SOLA-17-Quaternion} for an in-depth reference) may be represented by either of these equivalent forms,
\begin{align}
\begin{split}		
\bfq 
&= w+ix+jy+kz
=w+\bfv ~~ \in\bbH
\\
&
=\begin{bmatrix}w&x&y&z\end{bmatrix}\tr 
~\,=\begin{bmatrix}w\\\bfv\end{bmatrix} \quad~\, \in\bbH
~,
\end{split}
\end{align}
where $w,x,y,z\in\bbR$, and $i,j,k$ are three unit imaginary numbers such that $i^2=j^2=k^2=ijk=-1$.
The scalar $w$ is known as the scalar or real part, and $\bfv\in\bbH_p$ as the vector or imaginary part.
We note $\bbH_p$ the set of pure quaternions, \ie, of null scalar part, with dimension 3.
Inversion and composition are achieved by conjugation $\bfq\inv = \bfq^*$, where $\bfq^*\te w-\bfv$ is the conjugate, and product $\bfq_a\circ\bfq_b = \bfq_a\,\bfq_b$.

The group $\SO(3)$ is the group of special orthogonal matrices in 3D space, or rotation matrices, under matrix multiplication.
Inversion and composition are achieved with transposition and product as in all groups $\SO(n)$.

Both groups rotate 3-vectors. 
They have isomorphic tangent spaces whose elements are identifiable with rotation vectors in $\bbR^3$, so we study them together.
It is in this space $\bbR^3$ where we define the vectors of rotation rate $\bw\te\bfu\omega$, angle-axis $\bth\te\bfu\theta$, and all perturbations and uncertainties.

The quaternion manifold $S^3$ is a double cover of $\SO(3)$, \ie, $\bfq$ and $-\bfq$ represent the same rotation $\bfR$. 
The first cover corresponds to quaternions with positive real part $w>0$.
The two groups can be considered isomorphic up to the first cover.

\subsection{Exp and Log maps}

The Exp and Log maps may be defined for quaternions of $S^3$ and rotation matrices of $SO(3)$. 
For quaternions $\bfq=(w,\bfv)\in\bbH$ we have%
\if \examples y (see \exRef{ex:S3}), \else, \fi
\begin{align}
\bfq= \Exp(\theta\bfu) &\te \cos(\theta/2) + \bfu\sin(\theta/2) &&\in\bbH\\ 
\theta\bfu = \Log(\bfq) &\te 2\,\bfv\frac{\arctan({\norm{\bfv},w})}{\norm{\bfv}}&&\in\bbR^3
~.
\end{align}
We can avoid eventual problems due to the double cover of $\bfq$ by ensuring that its scalar part $w$ is positive before doing the $\Log$. 
If it is not, we can substitute $\bfq$ by $-\bfq$ before the $\Log$.

For rotation matrices we have%
\if \examples y (see \exRef{ex:SO3_exp}), \else, \fi
\begin{align}
\bfR= \Exp(\theta\bfu) &\te \bfI + \sin\theta\hatx{\bfu} + (1-\cos\theta)\hatx{\bfu}^2~ \label{equ:rodrigues} \in\bbR^{3\tcross3}\\ 
\theta\bfu = \Log(\bfR) &\te \frac{\theta(\bfR-\bfR\tr)^\vee}{2\sin\theta} \quad\in\bbR^3
~,
\end{align}
with $\theta=\cos\inv\big(\frac{\trace(\bfR)-1}{2}\big)$.

\subsection{Rotation action}

Given the expressions above for the quaternion and the rotation matrix, the rotation action of quaternions on 3-vectors is performed by the double quaternion product,
\begin{align}
\bfx' &= \bfq\,\bfx\,\bfq^* \\
\intertext{while rotation matrices use a single matrix product,}
\bfx' &= \bfR\bfx
~.
\end{align}
Both correspond to a right-hand rotation of $\theta$ rad around the axis $\bfu$.
Identifying in them $\bfx$ and $\bfx'$ yields the identity
\begin{align}\label{equ:q2R}
\bfR
(\bfq) \!=\!\! 
\begin{bsmallmatrix}
w^2+x^2-y^2-z^2 &~ 2(xy-wz) &~ 2(xz+wy) \\ 
2(xy+wz) &~ w^2-x^2+y^2-z^2 &~ 2(yz-wx) \\
2(xz-wy) &~ 2(yz+wx) &~ w^2-x^2-y^2+z^2
\end{bsmallmatrix}\!
\end{align}

\subsection{Elementary Jacobian blocks}

Since our defined derivatives map tangent vector spaces, and these spaces coincide for the 3D rotation manifolds of $S^3$ and $SO(3)$, \ie, $\bth=\Log(\bfq)=\Log(\bfR)$, it follows that the Jacobians are independent of the representation used ($\bfq$ or $\bfR$). 
We thus consider generic 3D rotation elements and note them with the sans-serif font $\sR$. 
%

\subsubsection{Adjoint}

We have from \eqRef{equ:Adj4}
\begin{align*}
\Ad{\sR} \bth
&= (\bfR\hatx{\bth}\bfR\tr)^\vee 
= (\hatx{(\bfR\bth)})^\vee 
= \bfR\bth
\end{align*}
therefore
\begin{align}
\Ad{\sR} = \bfR~,
\end{align}
which means, just to clarify it once again, that $\Ad[S^3]{\bfq}=\bfR(\bfq)$, see \eqRef{equ:q2R}, and $\Ad[\SO(3)]{\bfR}=\bfR$.

\subsubsection{Inversion, composition}
\label{sec:SO3_inv_comp}

We have from \secRef{sec:jacs_elementary},
\begin{align}
\mjac{\sR\inv}{\sR} &= -\Ad{\sR} ~= -\bfR \\
\mjac{\sQ\sR}{\sQ} &= \Ad{\sR}\inv = ~\bfR\tr \\
\mjac{\sQ\sR}{\sR} &= \bfI ~.
\end{align}

\subsubsection{Right and left Jacobians}

They admit the closed forms \cite[pag.~40]{CHIRIKJIAN-11}, 
\begin{align}
\mjac{}{r}(\bth) &= \bfI \!-\! \frac{1\!-\!\cos\theta}{\theta^2}\hatx{\bth} + \frac{\theta\!-\!\sin\theta}{\theta^3}\hatx{\bth}^2\\
\mjac{}{r}\inv(\bth) &= \bfI \!+\! \frac12\hatx{\bth} \!+\! \left(\frac{1}{\theta^2} \!-\! \frac{1\!+\!\cos\theta}{2\theta\sin\theta}\right)\hatx{\bth}^2 \\
\mjac{}{l}(\bth) &= \bfI + \frac{1-\cos\theta}{\theta^2}\hatx{\bth} + \frac{\theta-\sin\theta}{\theta^3}\hatx{\bth}^2 \label{equ:SO3_Jl} \\
\mjac{}{l}\inv(\bth) &= \bfI - \frac12\hatx{\bth} + \left(\frac1{\theta^2} - \frac{1+\cos\theta}{2\theta\sin\theta}\right)\hatx{\bth}^2 \label{equ:SO3_Jl_inv}
\end{align}
where we can observe that
\begin{align}
\mjac{}{l} &= \mjac{}{r}\tr 
~,
&
\mjac{}{l}\inv &= \mjac{}{r}^{-\top}
~.
\end{align}

\subsubsection{Right- plus and minus}

We have for $\bth=\sQ\om\sR$,
\begin{align}
\mjac{\sR\op\bth}{\sR}   
 &= \bfR(\bth)\tr 
 &
\mjac{\sR\op\bth}{\bth} 
 &= \mjac{}{r}(\bth)
 \\
\mjac{\sQ\om\sR}{\sQ} 
 &= \mjac{-1}{r}(\bth)
 &
\mjac{\sQ\om\sR}{\sR} 
 &= -\mjac{-1}{l}(\bth) 
\end{align}

\subsubsection{Rotation action} 
\label{sec:jac_SO3_action}

We have
\begin{align}
\mjac{\sR\cdot\bfv}{\sR} 
\small
&\te \lim_{\bth\to0}\frac{(\bfR\op\bth)\bfv-\bfR\bfv}{\bth} = \notag \\
\lim_{\bth\to0}\frac{\bfR\Exp(\bth)\bfv-\bfR\bfv}{\bth} 
&= \lim_{\bth\to0}\frac{\bfR(\bfI\!+\!\hatx{\bth})\bfv-\bfR\bfv}{\bth} \notag \\
= \lim_{\bth\to0}\frac{\bfR\hatx{\bth}\bfv}{\bth} 
&= \lim_{\bth\to0}\frac{-\bfR\hatx{\bfv}\bth}{\bth} 
= -\bfR\hatx{\bfv} 
\end{align}
where we used the properties $\Exp(\bth) \approx \bfI + \hatx{\bth}$ and $\hatx{\bfa}\bfb = -\hatx{\bfb}\bfa$. 
The second Jacobian yields,
\begin{align}
\mjac{\sR\cdot\bfv}{\bfv}
&\te \lim_{\partial\bfv\to0}\frac{\bfR(\bfv+\partial\bfv)-\bfR\bfv}{\partial\bfv} 
= \bfR
~.
\end{align}


\section{The 2D rigid motion group $SE(2)$}
\label{sec:SE2}

We write elements of the rigid motion group $\SE(2)$ as 
\begin{align}
\bfM= \begin{bmatrix}
\bfR & \bft \\ \bf0 & 1
\end{bmatrix} \in \SE(2) \subset \bbR^{3\times3}
~,
\end{align}
with $\bfR\in\SO(2)$ a rotation and $\bft\in\bbR^2$ a translation.
The Lie algebra and vector tangents are formed by elements of the type
\begin{align} 
\bftau^\wedge
  &= \begin{bmatrix}\hatx{\theta} & \bfrho \\ \bf0 & 0\end{bmatrix} \in \se(2)
  ~~,
& 
\bftau
  &= \begin{bmatrix}\bfrho \\ \theta\end{bmatrix}\in\bbR^3 
~.
\end{align}

\subsection{Inverse, composition}

Inversion and composition are performed respectively with matrix inversion and product,
\begin{align}
\bfM\inv &= \begin{bmatrix}
\bfR\tr & -\bfR\tr\bft \\ \bf0 & 1
\end{bmatrix} 
\\
\bfM_a\,\bfM_b &= \begin{bmatrix}
\bfR_a\bfR_b & \bft_a+\bfR_a\bft_b \\ \bf0 & 1
\end{bmatrix} 
~.
\end{align}

\subsection{Exp and Log maps}

Exp and Log are implemented via exponential maps directly from the scalar tangent space $\bbR^3\cong\se(2)=\mtan{\SE(2)}$ ---  see \cite{EADE-Lie} for the derivation,
\begin{align}
  \bfM = \Exp(\bftau) 
    &\te \begin{bmatrix}\Exp(\theta) & \bfV(\theta)\,\bfrho \\ \bf0 & 1  \end{bmatrix}  \label{equ:SE2_Exp} \\
  \bftau = \Log(\bfM) 
    &\te \begin{bmatrix} \bfV\inv(\theta)\, \bft \\ \Log(\bfR) \end{bmatrix}~.
\end{align}
with
\begin{align}
  \bfV(\theta)
  &= 
  \frac{\sin\theta}{\theta}\bfI + \frac{1-\cos\theta}{\theta}\hatx{1}
  ~.
\end{align}

\subsection{Jacobian blocks}
\label{sec:derivatives_SE2}

\subsubsection{Adjoint}

The adjoint is easily found from \eqRef{equ:Adj4} using the fact that planar rotations commute,
\begin{align*}
\Ad[\SE(2)]{\bfM} \bftau &= (\bfM \bftau^\wedge \bfM\inv)^\vee 
= \begin{bmatrix}
\bfR\bfrho-\hatx{\theta}\bft \\ \theta
\end{bmatrix} = \Ad[\SE(2)]{\bfM} \begin{bmatrix}
\bfrho\\\theta
\end{bmatrix}
~,
\end{align*}
leading to
\begin{align}
\Ad[\SE(2)]{\bfM} = \begin{bmatrix}
\bfR & -\hatx{1}\bft \\ \bf0 & 1
\end{bmatrix}
~.
\end{align}

\subsubsection{Inversion, composition}

We have from \secRef{sec:jacs_elementary},
\begin{align}
\mjac{\bfM\inv}{\bfM} &= -\Ad[\SE(2)]{\bfM} = \begin{bmatrix} -\bfR & \hatx{1}\bft \\ \bf0 & -1 \end{bmatrix}  \\
\mjac{\bfM_a\bfM_b}{\bfM_a} &= \Ad[\SE(2)]{\bfM_b}\inv ~= \begin{bmatrix} \bfR_b\tr & \bfR_b\tr\hatx{1}\bft_b \\ \bf0 & 1 \end{bmatrix} \\
\mjac{\bfM_a\bfM_b}{\bfM_b} &= \bfI 
~.
\end{align}

\subsubsection{Right and left Jacobians}

We have from \cite[pag.~36]{CHIRIKJIAN-11},
\begin{align}
\mjac{}{r} &= \begin{bsmallmatrix}
\sin\theta/\theta & (1-\cos\theta)/\theta & (\theta \rho_1 - \rho_2 + \rho_2 \cos\theta - \rho_1 \sin\theta)/\theta^2 \\
(\cos\theta-1)/\theta & \sin\theta/\theta & (\rho_1 + \theta \rho_2 - \rho_1 \cos\theta - \rho_2 \sin\theta)/\theta^2 \\
0 & 0 & 1
\end{bsmallmatrix} \\
\mjac{}{l} &= \begin{bsmallmatrix}
\sin\theta/\theta & (\cos\theta-1)/\theta & (\theta \rho_1 + \rho_2 - \rho_2 \cos\theta - \rho_1 \sin\theta)/\theta^2 \\
(1-\cos\theta)/\theta & \sin\theta/\theta & (-\rho_1 + \theta \rho_2 + \rho_1 \cos\theta - \rho_2 \sin\theta)/\theta^2 \\
0 & 0 & 1
\end{bsmallmatrix} 
~.
\end{align}

\subsubsection{Rigid motion action}

We have the action on points $\bfp$,
\begin{align}
\bfM\cdot\bfp &\te \bft+\bfR\bfp
~,
\end{align}
therefore and since for ${\bftau}\to0$ we have $\Exp(\bftau)\to\bfI+\bftau\hhat$, 
\begin{align}
\mjac{\bfM\cdot\bfp}{\bfM} 
  &= 
  \lim_{\bftau\to0}\frac{\bfM\Exp(\bftau)\cdot\bfp - \bfM\cdot\bfp}{\bftau}
  = 
  \begin{bmatrix}\bfR & \bfR\hatx{1}\bfp\end{bmatrix} \\
\mjac{\bfM\cdot\bfp}{\bfp} &= \bfR 
~.
\end{align}
%


\section{The 3D rigid motion group $SE(3)$}
\label{sec:SE3}

We write elements of the 3D rigid motion group $\SE(3)$ as 
\begin{align}
\bfM= \begin{bmatrix}
\bfR & \bft \\ \bf0 & 1
\end{bmatrix} \in \SE(3) \subset \bbR^{4\times4}
~,
\end{align}
with $\bfR\in\SO(3)$ a rotation matrix and $\bft\in\bbR^3$ a translation vector.
The Lie algebra and vector tangents are formed by elements of the type
\begin{align}
\bftau^\wedge
  &= \begin{bmatrix}\hatx{\bth} & \bfrho \\ \bf0 & 0\end{bmatrix} \in \se(3)
  ~~,
& 
\bftau
  &= \begin{bmatrix}\bfrho \\ \bth\end{bmatrix}\in\bbR^6 
~.
\end{align}

\subsection{Inverse, composition}

Inversion and composition are performed respectively with matrix inversion and product,
\begin{align}
\bfM\inv &= \begin{bmatrix}
\bfR\tr & -\bfR\tr\bft \\ \bf0 & 1
\end{bmatrix} 
\\
\bfM_a\,\bfM_b &= \begin{bmatrix}
\bfR_a\bfR_b & \bft_a+\bfR_a\bft_b \\ \bf0 & 1
\end{bmatrix} 
~.
\end{align}

\subsection{Exp and Log maps}

Exp and Log are implemented via exponential maps directly from the vector tangent space $\bbR^6\cong\se(3)=\mtan{\SE(3)}$ ---  see \cite{EADE-Lie} for the derivation,
\begin{align}
  \bfM = \Exp(\bftau) 
    &\te \begin{bmatrix}\Exp(\bth) & \bfV(\bth)\,\bfrho \\ \bf0 & 1  \end{bmatrix} \label{equ:SE3_Exp} \\
  \bftau = \Log(\bfM) 
    &\te \begin{bmatrix} \bfV\inv(\bth)\, \bft \\ \Log(\bfR) \end{bmatrix}~.
\end{align}
with (recall for $\Log(\bfM)$ that $\bth=\theta\bfu=\Log(\bfR)$)
\begin{align}
\bfV(\bth) = \bfI 
  + \frac{1-\cos\theta}{\theta^2}\hatx{\bftheta}
  + \frac{\theta-\sin\theta}{\theta^3}\hatx{\bftheta}^2
~
\end{align}
which, notice, matches \eqRef{equ:SO3_Jl} exactly.

\subsection{Jacobian blocks}
\label{sec:derivatives_SE3}

\subsubsection{Adjoint}

We have (see \exRef{ex:SE3_adjoint}),
\begin{align}
\Ad[\SE(3)]{\bfM}\bftau 
  &= (\bfM\bftau^\wedge\bfM\inv)^\vee \notag 
  = \begin{bmatrix}
  \bfR\bfrho+\hatx{\bft}\bfR\bth \\
  \bfR\bth
  \end{bmatrix} \notag 
  = \Ad[\SE(3)]{\bfM}\begin{bmatrix}
  \bfrho \\ \bth
  \end{bmatrix} \notag 
\end{align}
therefore,
\begin{align}
  \Ad[\SE(3)]{\bfM} &= \begin{bmatrix}
  \bfR & \hatx{\bft}\bfR \\ 0 & \bfR
  \end{bmatrix} \in \bbR^{6\times6}
  ~.
\end{align}

\subsubsection{Inversion, composition}

We have from \secRef{sec:jacs_elementary},
\begin{align}
\mjac{\bfM\inv}{\bfM} &= - \begin{bmatrix}
  \bfR & \hatx{\bft}\bfR \\ 0 & \bfR
  \end{bmatrix} \\
\mjac{\bfM_a\bfM_b}{\bfM_a} &=   \begin{bmatrix}
  \bfR_b\tr & -\bfR_b\tr\hatx{\bft_b} \\ 0 & \bfR_b\tr
  \end{bmatrix} \\
\mjac{\bfM_a\bfM_b}{\bfM_b} &= \bfI_6  
~.
\end{align}

\subsubsection{Right and left Jacobians}

Closed forms of the left Jacobian and its inverse are given by Barfoot in \cite{BARFOOT-14},
\begin{subequations}
\begin{align}
\mjac{}{l}(\bfrho,\bth) &= \begin{bsmallmatrix}
\mjac{}{l}(\bth) & \bfQ(\bfrho,\bth) \\
\bf0 & \mjac{}{l}(\bth)
\end{bsmallmatrix}
\\
\mjac{-1}{l}(\bfrho,\bth) &= \begin{bsmallmatrix}
\mjac{-1}{l}(\bth) & -\mjac{-1}{l}(\bth)\,\bfQ(\bfrho,\bth)\,\mjac{-1}{l}(\bth) \\
\bf0 & \mjac{-1}{l}(\bth)
\end{bsmallmatrix}
\end{align}
\end{subequations}
where $\mjac{}{l}(\bth)$ is the left Jacobian of $\SO(3)$, see \eqRef{equ:SO3_Jl}, and
\newcommand{\rhox}{\bfrho_\times}
\newcommand{\bthx}{\bth_\times}
\begin{align}
\bfQ(\bfrho,\bth) =& 
  \frac12\rhox 
  + \frac{\theta\!-\!\sin\theta}{\theta^3}(\bthx\rhox+\rhox\bthx+\bthx\rhox\bthx) 
  \notag\\
  &- \frac{1\! - \!\frac{\theta^2}{2}\! -\! \cos\theta}{\theta^4}(\bthx^2\rhox+\rhox\bthx^2-3\bthx\rhox\bthx)
  \notag\\
  &-\frac12\left(\frac{1 -  \frac{\theta^2}{2} - \cos\theta}{\theta^4} 
                  - 3\frac{\theta-\sin\theta-\frac{\theta^3}{6}}{\theta^5}\right)
  \notag\\
  &\times (\bthx\rhox\bthx^2 + \bthx^2\rhox\bthx)
~.
\end{align}
The right Jacobian and its inverse are obtained using \eqRef{equ:Jr_minus}, that is, $\mjac{}{r}(\bfrho,\bth)=\mjac{}{l}(-\bfrho,-\bth)$ and $\mjac{-1}{r}(\bfrho,\bth)=\mjac{-1}{l}(-\bfrho,-\bth)$. 

\subsubsection{Rigid motion action}
\label{sec:jac_SE3_action}
We have the action on points $\bfp$,
\begin{align}
\bfM\cdot\bfp &\te \bft+\bfR\bfp
~,
\end{align}
therefore and since for ${\bftau}\to0$ we have $\Exp(\bftau)\to\bfI+\bftau\hhat$, 
\begin{align}
\mjac{\bfM\cdot\bfp}{\bfM} 
  &= 
  \lim_{\bftau\to0}\frac{\bfM\Exp(\bftau)\cdot\bfp - \bfM\cdot\bfp}{\bftau}
  = \begin{bmatrix}\bfR & -\bfR\hatx{\bfp} \end{bmatrix} \\
\mjac{\bfM\cdot\bfp}{\bfp} &= \bfR
~.
\end{align}

\balance

\section{The translation groups $(\bbR^n,+)$ and $T(n)$}
\label{sec:Tn}

The group $(\bbR^n,+)$ is the group of vectors under addition and can be regarded as a translation group. We deem it \emph{trivial} in the sense that the group elements, the Lie algebra, and the tangent spaces are all the same, so $\bft=\bft^\wedge=\Exp(\bft)$.
Its equivalent matrix group (under multiplication) is the translation group $T(n)$, whose group, Lie algebra and tangent vector elements are,
\begin{align*}
\bfT &\te \begin{bmatrix}
\bfI & \bft \\ \bf0 & 1
\end{bmatrix} \in T(n),
&
\bft^\wedge &\te \begin{bmatrix}
\bf0 & \bft \\ \bf0 & 0
\end{bmatrix}\in\frak{t}(n),
&
\bft &\in\bbR^n
~.
\end{align*}
Equivalence is easily verified by observing that $\bfT(\bf0)=\bfI$, $\bfT(-\bft)=\bfT(\bft)\inv$, and that the commutative composition
\begin{align*}
\bfT_1\bfT_2 = \begin{bmatrix}
\bfI & \bft_1+\bft_2 \\ \bf0 & 1
\end{bmatrix}
~,
\end{align*}
effectively adds the vectors $\bft_1$ and $\bft_2$ together.
Since the sum in $\bbR^n$ is commutative, so is the composition product in $T(n)$.
Since $T(n)$ is a subgroup of $\SE(n)$ with $\bfR=\bfI$, we can easily determine its exponential map by taking \eqssRef{equ:SE2_Exp,equ:SE3_Exp} with $\bfR=\bfI$ and generalizing to any $n$, 
\begin{align}
\Exp&:&\bbR^n\to T(n)~; &&
\bfT &= \Exp(\bft) 
 =\begin{bmatrix}
  \bfI & \bft \\ \bf0 & 1
 \end{bmatrix}
 ~.
\intertext{The $T(n)$ exponential can be obtained also from the Taylor expansion of $\exp(\bft\hhat)$ noticing that $(\bft\hhat)^2=\bf0$.
This serves as immediate proof for the equivalent exponential of the $(\bbR^n,+)$ group, which is the identity,}
\Exp&:&\bbR^n\to \bbR^n &&
\bft &= \Exp(\bft)
~.
\end{align}
This derives in trivial, commutative, right- and left- alike, plus and minus operators in $\bbR^n$,
\begin{align}
\bft_1\op\bft_2   &= \bft_1+\bft_2 \\
\bft_2\ominus\bft_1 &= \bft_2-\bft_1
~.
\end{align}

\subsection{Jacobian blocks}

We express translations indistinctly for $T(n)$ and $\bbR^n$, and note them $\sS$ and $\sT$.
The Jacobians are trivial (compare them with those of $S^1$ and $\SO(2)$ in \secRef{sec:SO2_jacs}),
\begin{align}
\Ad{\sT}       
 &= ~\bfI 
 &&\in\bbR^{n\times n} \\
\mjac{\sT\inv}{\sT} 
 &= -\bfI 
 &&\in\bbR^{n\times n} \\
\mjac{\sT\circ\sS}{\sT} = ~\mjac{\sT\circ\sS}{\sS} 
 &= \bfI 
 &&\in\bbR^{n\times n} \\
\mjac{}{r} = ~\mjac{}{l} ~~
 &= \bfI 
 &&\in\bbR^{n\times n} \\
\mjac{\sT\op\bfv}{\sT} = ~~\mjac{\sT\op\bfv}{\bfv} 
 &= \bfI 
 &&\in\bbR^{n\times n} \\
\mjac{\sS\om\sT}{\sS} = -\mjac{\sS\om\sT}{\sT} 
 &= \bfI 
 &&\in\bbR^{n\times n} 
~.
\end{align}

\end{appendices}

\bibliographystyle{IEEEtran}
\bibliography{bibSLAM,diff_drive_bib}

\begin{thebibliography}{10}
\providecommand{\url}[1]{#1}
\csname url@samestyle\endcsname
\providecommand{\newblock}{\relax}
\providecommand{\bibinfo}[2]{#2}
\providecommand{\BIBentrySTDinterwordspacing}{\spaceskip=0pt\relax}
\providecommand{\BIBentryALTinterwordstretchfactor}{4}
\providecommand{\BIBentryALTinterwordspacing}{\spaceskip=\fontdimen2\font plus
\BIBentryALTinterwordstretchfactor\fontdimen3\font minus
  \fontdimen4\font\relax}
\providecommand{\BIBforeignlanguage}[2]{{%
\expandafter\ifx\csname l@#1\endcsname\relax
\typeout{** WARNING: IEEEtran.bst: No hyphenation pattern has been}%
\typeout{** loaded for the language `#1'. Using the pattern for}%
\typeout{** the default language instead.}%
\else
\language=\csname l@#1\endcsname
\fi
#2}}
\providecommand{\BIBdecl}{\relax}
\BIBdecl

\bibitem{ABBASPOUR-2007-Basic_Lie_theory}
\BIBentryALTinterwordspacing
H.~Abbaspour and M.~Moskowitz, \emph{Basic Lie Theory}.\hskip 1em plus 0.5em
  minus 0.4em\relax WORLD SCIENTIFIC, 2007. [Online]. Available:
  \url{https://worldscientific.com/doi/abs/10.1142/6462}
\BIBentrySTDinterwordspacing

\bibitem{Howe-Basic_Lie}
R.~Howe, ``Very basic {L}ie theory,'' \emph{The American Mathematical Monthly},
  vol.~90, pp. 600--623, 1983.

\bibitem{STILLWELL-08}
J.~Stillwell, \emph{Naive {L}ie Theory}.\hskip 1em plus 0.5em minus 0.4em\relax
  Springer-Verlag New York, 2008.

\bibitem{BARFOOT-17-Estimation}
T.~D. Barfoot, \emph{State Estimation for Robotics}.\hskip 1em plus 0.5em minus
  0.4em\relax Cambridge University Press, 2017.

\bibitem{EADE-Lie}
E.~Eade, ``Lie groups for 2d and 3d transformations,'' Tech. Rep.

\bibitem{forster2017-TRO}
C.~Forster, L.~Carlone, F.~Dellaert, and D.~Scaramuzza, ``On-manifold
  preintegration for real-time visual--inertial odometry,'' \emph{IEEE
  Transactions on Robotics}, vol.~33, no.~1, pp. 1--21, 2017.

\bibitem{DERAY-20-manif}
\BIBentryALTinterwordspacing
J.~Deray and J.~Sol{\`a}, ``Manif: A micro lie theory library for state
  estimation in robotics applications,'' \emph{Journal of Open Source
  Software}, vol.~5, no.~46, p. 1371, 2020. [Online]. Available:
  \url{https://doi.org/10.21105/joss.01371}
\BIBentrySTDinterwordspacing

\bibitem{SOLA-17-Quaternion}
\BIBentryALTinterwordspacing
J.~Sol{\`{a}}, ``Quaternion kinematics for the error-state {K}alman filter,''
  \emph{CoRR}, vol. abs/1711.02508, 2017. [Online]. Available:
  \url{http://arxiv.org/abs/1711.02508}
\BIBentrySTDinterwordspacing

\bibitem{GALLEGO-13}
G.~Gallego and A.~Yezzi, ``A compact formula for the derivative of a 3-{D}
  rotation in exponential coordinates,'' Tech. Rep., 2013.

\bibitem{BARFOOT-14}
T.~D. Barfoot and P.~T. Furgale, ``Associating uncertainty with
  three-dimensional poses for use in estimation problems,'' \emph{IEEE
  Transactions on Robotics}, vol.~30, no.~3, pp. 679--693, June 2014.

\bibitem{CHIRIKJIAN-11}
\BIBentryALTinterwordspacing
G.~Chirikjian, \emph{Stochastic Models, Information Theory, and Lie Groups,
  Volume 2: Analytic Methods and Modern Applications}, ser. Applied and
  Numerical Harmonic Analysis.\hskip 1em plus 0.5em minus 0.4em\relax
  Birkh{\"a}user Boston, 2011. [Online]. Available:
  \url{https://books.google.ch/books?id=hZ1CAAAAQBAJ}
\BIBentrySTDinterwordspacing

\bibitem{DELLAERT-IJRR-06}
F.~Dellaert and M.~Kaess, ``{Square Root SAM: Simultaneous Localization and
  Mapping via Square Root Information Smoothing},'' vol.~25, no.~12, pp.
  1181--1203, 2006.

\bibitem{KAESS-11-ISAM2}
M.~Kaess, H.~Johannsson, R.~Roberts, V.~Ila, J.~Leonard, and F.~Dellaert,
  ``{iSAM2}: {I}ncremental smoothing and mapping with fluid relinearization and
  incremental variable reordering,'' in \emph{Robotics and Automation (ICRA),
  2011 IEEE International Conference on}, May 2011, pp. 3281--3288.

\bibitem{KUMMERLE-11-G2O}
R.~Kummerle, G.~Grisetti, H.~Strasdat, K.~Konolige, and W.~Burgard, ``G2o: A
  general framework for graph optimization,'' in \emph{Robotics and Automation
  (ICRA), 2011 IEEE International Conference on}, May 2011, pp. 3607--3613.

\bibitem{ILA-17_SLAM++}
\BIBentryALTinterwordspacing
V.~Ila, L.~Polok, M.~Solony, and P.~Svoboda, ``{SLAM++} - a highly efficient
  and temporally scalable incremental {SLAM} framework,'' \emph{The
  International Journal of Robotics Research}, vol.~36, no.~2, pp. 210--230,
  2017. [Online]. Available: \url{https://doi.org/10.1177/0278364917691110}
\BIBentrySTDinterwordspacing

\end{thebibliography}

\end{document}